\documentclass{article}

 \usepackage[main, preprint]{neurips_2025}

\usepackage[utf8]{inputenc} % allow utf-8 input
\usepackage[T1]{fontenc}    % use 8-bit T1 fonts
\usepackage{hyperref}       % hyperlinks
\usepackage{url}            % simple URL typesetting
\usepackage{booktabs}       % professional-quality tables
\usepackage{amsfonts}       % blackboard math symbols
\usepackage{nicefrac}       % compact symbols for 1/2, etc.
\usepackage{microtype}      % microtypography
\usepackage{xcolor}         % colors
\usepackage{amsmath}
\usepackage{amssymb}
\usepackage{mathtools}
\usepackage{amsthm}
\usepackage{bm}
\usepackage{caption}
\usepackage{subcaption}
\usepackage{microtype}
\usepackage{graphicx}
\usepackage{booktabs}
\usepackage[capitalize,noabbrev]{cleveref}
\usepackage{diagbox}
\theoremstyle{plain}

\theoremstyle{definition}

\theoremstyle{remark}

\usepackage{multicol}
\usepackage{multirow}
\usepackage[disable,textsize=tiny]{todonotes}
\usepackage{algorithm}
\usepackage{algorithmic}
\title{Tokenizing Electron Cloud in Protein-Ligand Interaction Learning}

\author{
  Haitao Lin$^{1,2}$, \quad Odin Zhang$^{2}$, \quad Jia Xu$^1$, \quad Yunfan Liu$^1$, \\
  \textbf{  Zheng Cheng$^3$, \quad Lirong Wu$^1$,\quad  Yufei Huang$^1$, \quad Zhifeng Gao$^3$, \quad Stan Z. Li$^{1, *}$}  \\
  $^1$AI Lab, Research Center for Industries of the Future, Westlake University; \\
    $^2$Zhejiang University; \quad $^3$DP Technology; \quad $^{*}$Corresponding Author; \\
   \texttt{linhaitao@westlake.edu.cn} ;  \texttt{stan.zq.li@westlake.edu.cn} }

\begin{document}

\maketitle

\begin{abstract}
The affinity and specificity of protein-molecule binding directly impact functional outcomes, uncovering the mechanisms underlying biological regulation and signal transduction. Most deep-learning-based prediction approaches focus on structures of atoms or fragments. However, quantum chemical properties, such as electronic structures, are the key to unveiling interaction patterns but remain largely underexplored. To bridge this gap, we propose ECBind, a method for tokenizing electron cloud signals into quantized embeddings, enabling their integration into downstream tasks such as binding affinity prediction. By incorporating electron densities, ECBind helps uncover binding modes that cannot be fully represented by atom-level models. Specifically, \textit{to remove the redundancy inherent in electron cloud signals}, a structure-aware transformer and hierarchical codebooks encode 3D binding sites enriched with electron structures into tokens. These tokenized codes are then used for specific tasks with labels. 
\textit{To extend its applicability to a wider range of scenarios}, we utilize knowledge distillation to develop an electron-cloud-agnostic prediction model. Experimentally, ECBind demonstrates state-of-the-art performance across multiple tasks, achieving improvements of 6.42\% and 15.58\% in per-structure Pearson and Spearman correlation coefficients, respectively.\end{abstract}
\section{Introduction}
\label{sec:intro}
Protein-ligand binding interaction is important for understanding the mechanisms of biological regulation, providing a theoretical basis for the design and discovery of new drugs~\citep{Fu2018InsightsIT}. The binding process is characterized by two key attributes: (i) specificity, which enables the precise recognition of a highly specific binding partner over less specific alternatives; and (ii) affinity, which ensures that even a high concentration of weakly interacting partners cannot substitute for the effects of a low concentration of the specific partner with strong binding affinity~\citep{Du2016InsightsIP}. In the field of artificial-intelligence-aided drug design (AIDD), the learning of interaction patterns thus can be summarized into two tasks: (i) binary classification of active and inactive states of the binding complexes, and (ii) regression prediction of binding affinity.

In recent years, the abundance of protein-molecule binding structure data, coupled with the rise of 3D graph neural networks~(GNNs), has significantly accelerated progress in the tasks~\citep{egnn, messagepassing,Klicpera2020DirectionalMP}.
Most of these approaches~\citep{Klicpera2021GemNetUD,schnet,infomaxgnn} treat the atoms or amino acids as fundamental units, modeling the binding graph at different scales (e.g., atom-level or fragment-level). By constructing translational, rotational, and reflectional equivariant or invariant features, they leverage the neural networks to extract geometric representations effectively and show promising performance in protein-ligand interaction tasks~\citep{egnn,e3nn}. 

One limitation is that their finest granularity is typically at the atomic level. However, at the quantum scale, electronic structures, compared to atoms within a molecule, reflect the fundamental forces governing intermolecular interactions~\citep{bioforces}. Specifically, beyond the 3D atomic structures represented by the ball-and-stick descriptions, which struggle to accurately capture interactions such as electrostatic forces, van der Waals interactions, hydrogen bonding, and $\pi$-$\pi$ stacking, electron cloud provides a detailed spatial distribution of electrons. It reveals critical features, such as charge polarization and lone pairs which are essential for understanding interaction patterns like non-covalent interactions and molecular polarity~\citep{revealing_non-covalent}.
Moreover, in identifying binding sites, the shape and polarity of the electron cloud in molecules play a significant role in determining their affinity and selectivity for targets, as the polar regions of a drug molecule often complement those of the receptor, enhancing binding specificity~\citep{ecloud_binding}.
However, while perceiving electron clouds is crucial, integrating such descriptions of electronic structures into computational models poses significant challenges. This is due to \textit{(i) their inherent redundancy as natural signals} and \textit{(ii) difficulty accessibility resulting from costly computational complexity}~\citep{ecloud_complexity}.

To bridge the gap, we introduce \textbf{ECBind}, a pretraining and finetuning paradigm designed to integrate protein-ligand electronic and atomic structures into an interaction prediction model. The pretraining phase enables the model to perceive both \underline{e}lectron \underline{c}loud as electronic structural descriptors and the full-atom structures of \underline{bind}ing proteins and ligands, while the finetuning process allows the model to focus on specific tasks. Besides, to avoid the heavy computational cost in computing electron cloud, we develop a lightweight electron-cloud-agnostic student model with knowledge distillation. Empirically, ECBind demonstrates state-of-the-art performance across multiple tasks, achieving improvements of 6.42\% and 15.58\% in per-structure Pearson and Spearman correlation coefficients on relative binding affinity prediction, with little performance degeneration in the student model.

\section{Background}
\textbf{3D atomic binding graph.} We define the binding system composed of a protein-ligand pair as $\mathcal{B} = (\bm{V}, \bm{E}) = (\mathcal{P}, \mathcal{M})$, in which $\mathcal{P}$ is the protein and $\mathcal{M}$ is the molecule ligand. The node set of $\mathcal{P}$ is $\bm{V}_{\mathrm{rec}}$, containing $N_{\mathrm{rec}}$ atoms of proteins and $\bm{V}_{\mathrm{lig}}$ contains $N_{\mathrm{lig}}$ atoms of molecules. We denote the index set of the protein's atom as $\mathcal{I}_{\mathrm{rec}}$ and the molecule's atoms as $\mathcal{I}_{\mathrm{lig}}$, with $N_{\mathrm{fa}} = N_{\mathrm{rec}} + N_{\mathrm{lig}}$, leading to $\bm{V}_{\mathrm{rec}} = \{(a_i, \bm{\mathrm{x}}^{\mathsf{a}}_i)\}_{i\in \mathcal{I}_{\mathrm{rec}}}$ and $\bm{V}_{\mathrm{lig}} = \{(a_i, \bm{\mathrm{x}}^{\mathsf{a}}_i)\}_{i\in \mathcal{I}_{\mathrm{lig}}}$. Here, $a_i$ is the atom types with $a_i = 1,\ldots, M$, and $\bm{\mathrm{x}}^{\mathsf{a}}_i \in \mathbb{R}^3$ is the corresponding 3D position. $\bm{E}$ is the edge set. 

\textbf{Electron cloud density.}  By employing numerical methods such as density functional theory~\citep{dft} or semi-empirical methods~\citep{xtb}, we can extract electron clouds from full atomic structures (including hydrogen atoms) as density maps of electronic structures. Typically, electron cloud is represented as $\mathcal{D} = \{(u_i, \bm{\mathrm{x}}^\mathsf{u}_i)\}_{1 \leq i \leq N_{\mathrm{ec}}}$, where $u_i$ represents the electron density at position $\bm{\mathrm{x}}^{\mathsf{u}}_i$. Since the electron cloud signals are usually distributed on 3D grids, high-resolution signals inevitably result in $N_{\text{ec}} \gg N_{\text{fa}}$. Moreover, while the dense electron cloud contains sufficient interaction patterns compared to atom-level signals, it is excessively redundant. These lead to the following challenges: 
\begin{itemize}
    \item  From a computational perspective, if we treat $|\mathcal{D}| = N_{\text{ec}}$ as the number of nodes in the graph, message passing becomes infeasible due to the overwhelming computational cost. The reason is that usually, the scales of a binding site are about $\sim 20$\AA, and if we set the resolution as 0.5 \AA, it will leads $(20/0.5)^3$ nodes.).
    \item From a signal processing perspective, such redundant signals carry limited semantic information and require further compression. Through compression can the resulting high-dimensional representations generalize more effectively and robustly to downstream tasks.
\end{itemize}

\textbf{Interaction prediction.} 
For the two attributes introduced in Sec.~\ref{sec:intro}, we define the specificity prediction as finding a mapping $f$ such that $\hat{y} = f(\mathcal{B}, \mathcal{D}) \in [0,1]$, representing the probability of the ligand molecule's activeness on the protein. Additionally, for the absolute binding affinity prediction, $\hat{y}_p = f(\mathcal{B}_p, \mathcal{D}_p) \in \mathbb{R}^+$ represents the binding strength in positive real values. For the relative binding affinity prediction, $\hat{y}_{pq} = f(\mathcal{B}_p, \mathcal{D}_p, \mathcal{B}_q, \mathcal{D}_q) \in \mathbb{R}$ indicates the difference in binding affinity as $\mathcal{B}_q = (\mathcal{P}_q, \mathcal{M}_q)$ changes to $\mathcal{B}_p = (\mathcal{P}_q, \mathcal{M}_p)$, with the corresponding changes in the electron cloud from $\mathcal{D}_q$ to $\mathcal{D}_p$. The encoder and decoder are all transformer-based, which can be regarded as graph neural networks for fully connected graphs.
\vspace{-0.5em}
\section{Method}
\subsection{Tokenizing Electron Cloud}\label{sec:ectoken}
\begin{figure*}[t]
    \centering
        \includegraphics[width=1.0\textwidth, trim=45 230 280 30, clip]{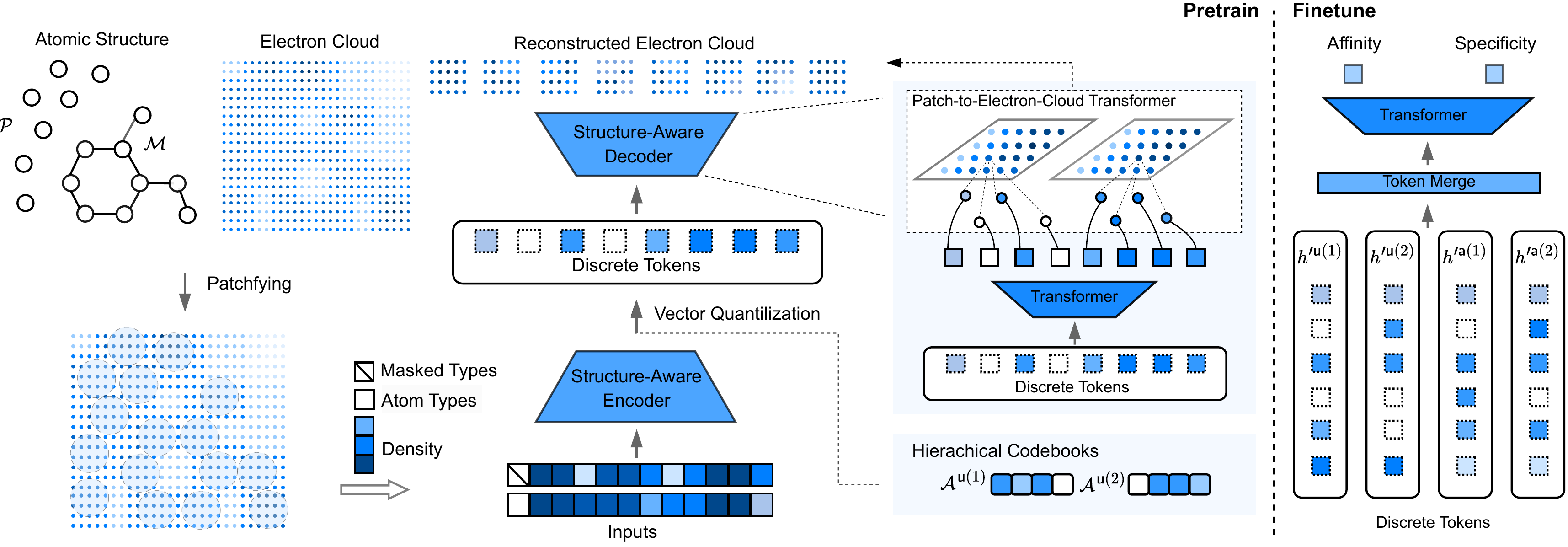} \vspace{-1em}
        \caption{Pretraining electron-cloud-aware codebooks and finetuning on the specific tasks. } \label{fig:ecworkflow}\vspace{-1em}
\end{figure*}

\vspace{-0.5em}
In order to (i) enable the learned embeddings to capture the electronic structure of the protein-ligand binding, (ii) reduce the redundancy in electron cloud signals generated by numerical methods, and (iii) enhance the generalizability of the learned representations for downstream tasks, we propose to pretrain an encoder with vector quantization as signal compression techniques to generate tokenized embeddings of the binding graph. Figure.~\ref{fig:ecworkflow} gives the workflows.
\subsubsection{Encoder architecture}
\label{sec:encoder}
\paragraph{Patchifying electron clouds.} 
We first use linear projectors to lift $u_i$ into a higher-dimensional space, denoted as $\bm{z}^{\mathsf{u}}_i$, and similarly, $a_i$ is lifted as $\bm{z}^{\mathsf{a}}_i$. To downsample the resolution of the signal $\{u_i\}_{1\leq i \leq N_\mathrm{ec}}$, we patchify the electron cloud signals, where each patch contains $K$ electron densities, with its center being the nearest atom position. As a result, we obtain $N_\mathrm{fa}$ patches, whose embedding sequence is represented as $\bm{Z}^{\mathsf{u}}_i = (\bm{z}^{\mathsf{a}}_i, \bm{z}^{\mathsf{u}}_{j_1}, \ldots, \bm{z}^{\mathsf{u}}_{j_K})$, and the corresponding position sequence as $\bm{X}^{\mathsf{u}}_i = (\bm{\mathrm{x}}^{\mathsf{a}}_i, \bm{\mathrm{x}}^{\mathsf{u}}_{j_1}, \ldots, \bm{\mathrm{x}}^{\mathsf{u}}_{j_K})$. Here, ${j_k}$ is the subsequence of $(j)_{j=1,\ldots,N_{\mathrm{ec}}}$, such that $j_k$ lies in the neighborhood of $i$, determined by the Euclidean distance $||\bm{\mathrm{x}}^{\mathsf{u}}_{j_k} - \bm{\mathrm{x}}^{\mathsf{a}}_{i}||_2$. 
However, in practice, $K$ cannot be set as a sufficiently large value to cover the entire electron cloud signal set. To address it, we employ two techniques to allow the model to perceive electron densities farther from the patch center as part of the input:
\begin{itemize}
    \item \textit{Random Selection as Input Augmentation}. Inspired by neural operators for spatially irregular inputs~\citep{clcrn, gkno}, we randomly sample points from the input grid, making the signals irregularly distributed to serve as inputs. Specifically, we use $r_\mathrm{in}$ to determine whether a given density value is included as part of the input. 
    \item \textit{Random Selection of Patch Elements}. Another technique involves setting a large $K_{\max}$ for the neighborhood of each $i$-th atom and then randomly selecting $K$ electron cloud signals from the $K_{\max}$ neighbors. The selection is performed with a probability $p_{j_k} = \frac{\exp(\|\bm{\mathrm{x}}^{\mathsf{u}}_{j_k} - \bm{\mathrm{x}}^{\mathsf{a}}_{i}\|_2/\gamma_\mathrm{pt})}{\sum_{k=1}^{K_{\max}}\exp(\|\bm{\mathrm{x}}^{\mathsf{u}}_{j_k} - \bm{\mathrm{x}}^{\mathsf{a}}_{i}\|_2/\gamma_\mathrm{pt})}$, where $\gamma_\mathrm{pt}$ is the predefined smoothing coefficient, set as 0.1. \vspace{-0.5em}
\end{itemize}
\textbf{Structure-aware transformer.}
We refer to GeoMHA from ESM3~\citep{esm3} as the structure-aware module to construct the encoder’s network architecture. However, while ESM3 operates at the residue level, we adapt it to the atom/electron cloud level. Specifically, for $\bm{Z} \in \mathbb{R}^{L\times D}$ representing the embedding of one patch with length $L$, and $\bm{X} \in \mathbb{R}^{L\times 3}$ denoting the corresponding positions (where $D$ is the embedding dimension), a linear layer is first applied to project $\bm{Z}$ into invariant-feature keys and queries, geometric-feature keys and queries, and values, denoted as $\bm{Q}^{\mathsf{iv}}, \bm{Q}^{\mathsf{ge}}, \bm{K}^{\mathsf{iv}}, \bm{K}^{\mathsf{ge}}, \bm{V} \in \mathbb{R}^{L\times H\times 3}$, where $H$ is the number of attention heads.
The first attention matrix, $\bm{\mathrm{A}}^{\mathsf{iv}} \in \mathbb{R}^{L\times L \times H}$, regulates the classical attention mechanism, while the second, $\bm{\mathrm{A}}^{\mathsf{ge}} \in \mathbb{R}^{L\times L \times H}$, incorporates geometric-awareness. Their difference collectively forms the final attention weights, $\bm{\mathrm{A}}$, enabling message-passing among $L$ input points, given by:
\begin{align}
\bm{\mathrm{A}} &= \mathrm{softmax}(\bm{\mathrm{A}}^{\mathsf{iv}} - \bm{\mathrm{A}}^{\mathsf{ge}})
\end{align}
$$
\text{where}\quad
\bm{\mathrm{A}}^{\mathsf{iv}} = \frac{\mathrm{softplus}(w^{\mathsf{iv}})}{\sqrt{3}}(\bm{Q}^{\mathsf{iv}})^{\intercal} \bm{K}^{\mathsf{iv}},\quad  \text{and} \quad 
\bm{\mathrm{A}}^{\mathsf{ge}} = \frac{\mathrm{softplus}(w^{\mathsf{ge}})}{\sqrt{3}} \mathrm{pdist}( \bm{X} \cdot \bm{Q}^{\mathsf{ge}}, \bm{X} \cdot \bm{K}^{\mathsf{ge}} ). \notag
$$
 $w^{\mathsf{iv}}$ and  $w^{\mathsf{ge}}$ are learnable scalars, and $\mathrm{pdist}$ is the pairwise-distance function. The output is computed as $\bm{O} = \bm{\mathrm{A}} \bm{V}$. Finally, a feed-forward network (FFN) with LayerNorm and residual connections projects $\bm{O}$ back into the space of $\bm{Z}$. Appendix.~\ref{app:invproof} gives a brief proof of the trans-rotational invariance of the extracted embedding $\bm{Z}$.

In specific, for a patch defined by $\bm{Z}^\mathsf{u}_i$ and $\bm{X}^\mathsf{u}_i$, the patch length is given by $L = K + 1$. After stacking multiple GeoMHA-based transformer layers to form the encoder $\phi^{\mathsf{ec}}(\cdot, \cdot)$, the representation of the patch center (indexed by 1) is extracted as the final representation of the patch, denoted as $\bm{h}^\mathsf{u}_i = [\phi^{\mathsf{ec}}(\bm{Z}^\mathsf{u}_i, \bm{X}^\mathsf{u}_i)]_1$.

\subsubsection{Codebook construction} The patchwise representation of $\bm{h}^\mathsf{u}_i$ will then be tokenized into quantized codes. Following recent advances in language modeling as different modalities require specific prompts, hierarchical codebook is used and decomposed into two types: 3D electronic structures and 2D atomic attributes, denoted as $\mathcal{A}^\mathsf{u} = \mathcal{A}^{\mathsf{u}(1)} \times \mathcal{A}^{\mathsf{u}(2)}$. Each $\mathcal{A}^{\mathsf{u}(k)}$ is parameterized with learnable embeddings $\{\bm{e}^{\mathsf{u}(k)}_j\}_{j=1}^{|\mathcal{A}^{\mathsf{u}(k)}|}$.

To generate the tokenized codes for different levels, $\bm{h}^\mathsf{u}_i$ is first split as $\bm{h}^\mathsf{u}_i = [\bm{h}^{\mathsf{u}(1)}_i; \bm{h}^{\mathsf{u}(2)}_i]$. Then, $\bm{h}^{\mathsf{u}(k)}_i$ is tokenized into discrete codes $\bm{h}_i'^{\mathsf{u}(k)}$ using vector quantization, as 
\begin{equation}
    \begin{aligned}
        \bm{h}_i'^{\mathsf{u}(k)} = \arg \min_j \|\bm{h}_i^{\mathsf{u}(k)} - \bm{e}_j^{\mathsf{u}(k)}\|_2, \quad  k = 1,2. \label{eq:vqu}
    \end{aligned}
\end{equation}

In practice, we employ spherical projection to regularize $\bm{h}_i^{\mathsf{u}(k)}$ and $\bm{e}_j^{\mathsf{u}(k)}$, which enhances training stability~\citep{huh2023improvedvqste}. Additionally, Repeated K-means~\citep{Chiu2022SelfsupervisedLW} is applied to mitigate codebook collapse.
\subsubsection{Structure reconstruction}
\label{sec:ecrec}
\textbf{Electron cloud reconstruction.}
By compressing $N_{\mathrm{ec}}$ signals into $N_{\mathrm{fa}}$ tokens, we aim for $\bm{H}'^{\mathsf{u}(1)} = \{\bm{h}'^{ \mathsf{u}(1)}_i\}_{1 \leq i \leq N_{\mathrm{fa}}}$ to retain key electronic structure information. Therefore, to pretrain the tokenizer composed of encoder $\phi^{\mathsf{ec}}$ and codebook $\mathcal{A}^{\mathsf{u}(1)}$, we hope to reconstruct the $N_\mathrm{ec}$ electron cloud signals using only $N_\mathrm{fa}$ patch-level embeddings. In detail, we first use layers of classical transformers for patch-level message passing, as
$\bm{H}^{\mathsf{pt} (1)} = \psi^\mathsf{pt}(\bm{H}'^{\mathsf{u}(1)})$. Then the patch-to-electron message-passing can be performed similarly to the structure-aware transformer in Sec.~\ref{sec:encoder}, with 
\begin{equation}
    \begin{aligned}
        \quad \quad \quad \bm{h}_j^{\mathsf{pe}} &= [\psi^{\mathsf{pe}}(\bm{H}^{\mathsf{pe}(1)}_j, \bm{{X}}^{\mathsf{pe}}_j)]_1, \quad \quad \\
        \text{where} \quad \bm{H}^{\mathsf{pe}(1)}_j = (\bm{0}, \bm{h}_{i_1}^{\mathsf{pt}(1)}, &\ldots, \bm{h}_{i_K}^{\mathsf{pt} (1)} ), \quad \quad  \text{ and} \quad 
        \bm{X}^{\mathsf{pe}}_j = (\bm{\mathrm{x}}^{\mathsf{u}}_{j},\bm{\mathrm{x}}_{i_1}^{\mathsf{pt}}, \ldots, \bm{\mathrm{x}}_{i_K}^{\mathsf{pt}} ) .
    \end{aligned}
\end{equation}
$\bm{0}$ is a zero vector like [CLS] in language modeling, $(i_k)_{1\leq k\leq K}$ is the subsequence of $(i)_{1\leq i \leq N_{\mathrm{fa}}}$ and forms the $K$-nearest neighborhood of $j$ according to their distance.  The patch-to-electron module will generate $N_\mathrm{ec}$ embeddings of $\{\bm{h}^{\mathsf{pe}}_j\}_{1\leq j \leq N_\mathrm{ec}}$. And the final embeddings of the electron cloud will be computed with another classical transformer $\bm{H}^{\mathsf{ec}} = \psi^{\mathsf{ec}}(\{\bm{h}^{\mathsf{pe}}_j\}_{1\leq j \leq N_\mathrm{ec}})\in \mathbb{R}^{N_{\mathrm{ec}} \times D}$, allowing electron-level message-passing. A linear projector is used to map $\bm{h}_j^{\mathsf{ec}}$ to $\hat{u}_j$ and Mean Square Error (MSE)  is employed to reconstruct the electron clouds, 
\begin{equation}
    L^{\mathsf{ec}} = \mathrm{MSE}_j (\hat{u}_j, u_j ), \quad \quad  \text{where } 1\leq j \leq N_{\mathrm{ec}}.
\end{equation}
$\mathrm{MSE}_j(\cdot, \cdot)$ indicates mean reduction across the index $j$.

\textbf{Masked atom type prediction.} Moreover, to enable the encoder to perceive the 2D graph topology, we use $\bm{h}'^{\mathsf{u}(2)}_i$, obtained from the codebook $\mathcal{A}^{\mathsf{u}(2)}$, as the embedding of 2D atomic attributes. Since each patch center is defined as the nearest atom position, $\bm{h}'^{\mathsf{u}(2)}_i$ can naturally be regarded as the embedding of the $i$-th atom. To train the tokenized representation effectively, we adopt techniques from Masked AutoEncoder (MAE)~\citep{mae}. Specifically, a ratio of atom types in the input is randomly selected and masked (Masking Ratio is set as 0.1), and the tokenizer extracts $\bm{h}'^{\mathsf{u}(2)}_i$ from the randomly masked inputs. Subsequently, an atom-level transformer is applied to decode $\bm{h}'^{\mathsf{u}(2)}_i$ into $\mathrm{logit}_i^{\mathsf{at}} \in \mathbb{R}^M$, as the predicted logits of the masked atom types' probabilities with cross-entropy (CE) loss utilized as
\begin{equation}
    L^{\mathsf{at}} = \text{CE}_i(\mathrm{logit}_i^{\mathsf{at}}, a_i), \quad   1\leq i \leq N_{\mathrm{fa}} \text{ and } a_i \text{ is masked}.
\end{equation}
Since the encoder $\phi^{\mathsf{ec}}$ are shared by the two tasks, we can regard this prediction task as using the contextual of unmasked atom types with the positions and the electron cloud shape to predict the masked atom types correctly.
\subsection{Tokenizing Atom-level Structure}\label{sec:fatoken}

 The electronic structure usually provides sophisticated information for local interaction, and besides it, we aim for our model to perceive a more global structural one. To achieve this, we establish an atom-level tokenizer that encodes each atom with its macro-environment~\citep{mapeppi}, enabling the model to capture long-range inter-atomic interactions. We use a full-atom GeoMHA-based transformer to encode the macro-environment~(See Appendix.~\ref{app:structtoken}), and a classical transformer to decode the tokenized representation into a structural one~(See Appendix.~\ref{app:structdec}). Two types of loss functions are used: (i) intra-molecular binned distance classification loss and (ii) inter-molecular interaction loss.~(See Appendix.~\ref{app:structdec})

\subsection{Training, Usage, and Distillation}
\textbf{Training.} 
We employ a pretrain-finetune paradigm to train our proposed ECBind. For the electron cloud tokenizer pretraining, besides the discussed loss, a commitment loss is also employed:
\begin{equation}
    L^{\mathsf{cmt}(k)} = \text{MSE}_i(\text{sg}(\bm{h}^{\mathsf{u}(k)}_i), \bm{e}_i^{\mathsf{u}(k)}),
\end{equation}
where the update for $\bm{e}_j^{\mathsf{u}(k)}$ is performed using an Exponential Moving Average (EMA), and $\text{sg}(\cdot)$ denotes the stop-gradient operation~\citep{bengio2013estimatingpropagatinggradientsstochastic}. 
The training objective for the tokenizer is:
\begin{equation}
    L^{\mathsf{u}} = L^{\mathsf{ec}} + L^{\mathsf{at}} + \alpha^{\mathsf{u}} (L^{\mathsf{cmt}(1)} + L^{\mathsf{cmt}(2)}),
\end{equation}
where $\alpha^{\mathsf{u}}$ is the loss weight, typically set to 10.0. The atom-level tokenizer training follows the same.

\textbf{Usage.} 
For downstream tasks such as binding affinity prediction, we employ an attention mechanism to adaptively learn the combination of codebooks as the representation:
\begin{equation}
    \begin{aligned}
        \bm{h}_i = \sum_\mathsf{s}\alpha_i^{\mathsf{s}} &\bm{h}_i'{^{\mathsf{s}}};\quad\quad \alpha_i^{\mathsf{s}} = \varphi(\{\bm{h}_i'{^{\mathsf{s}}}\}),\\
        &{\mathsf{s}\in\{\mathsf{u}(1),\mathsf{u}(2), \mathsf{a}(1), \mathsf{a}(2)\}}. \\
    \end{aligned}
\end{equation}
A lightweight network $\psi(\cdot)$ is subsequently stacked to decode $\bm{H} = \{\bm{h}_i\}_{1\leq i \leq N_\text{fa}}$ into the target value with task-specific objectives. For example, for absolute binding affinity prediction, the $p$-th complex structure generates the binding affinity ${y}_p$ as ground truth,  and $\hat{y}_p = \psi(\bm{H}_p) \in \mathbb{R}$ is fine-tuned with the objective $L^{\mathsf{ba}} = \text{MSE}(\hat{y}_p, y_p)$  ; For relative affinity prediction, the label is ${y}_{pq} = y_p - y_q$, and $\hat{y}_{pq} = \psi([\bm{H}_p, \bm{H}_q])$ is trained with
        $L^{\mathsf{ba}} = \text{MSE}_i(\hat{y}_{pq}, y_{pq}).$

\textbf{Distillation.} 
However, obtaining electron cloud signals is computationally expensive. To address this, we employ a student network, sharing the same architecture as the trained teacher, to distill knowledge~\citep{distill1, distill2, distill3}. In detail, the student approximates the merged tokens, avoiding the redundancy of electron cloud signals and reducing error accumulation. Instead of building an electron cloud approximator, we use only the full-atom structure of the binding site as input, with tokenized embeddings trained to approximate $\bm{h}_i$.
Notably, we find that the electron-cloud-agnostic student network achieves comparable performance (see Sec.~\ref{sec:exp}).
\section{Related Work}
\textbf{Tokenizing biomolecules.} The success of quantization techniques for image generation and representation~\citep{vqgan1,vqgan2, maskgit} boosts works of tokenizing biomolecules. Foldseek is motivated by expanding the codebook of 20 amino acids for fast retrieving protein data~\citep{foldseek}. 
Most existing methods tokenize atoms or fragments (e.g., amino acids) with their surrounding environments, enabling models to perceive molecular structure and topology~\citep{su2024saprot,protoken1,protoken2,moltoken1,moltoken2} for downstream tasks such as property prediction or structure generation~\citep{mapeppi,promptddg}. Recently, ESM3~\citep{esm3}, as a multimodal language model, generates tokens for reasoning over the sequence, structure, and function of proteins.
Alternatively, methods focusing on finer-grained representations, such as surface point clouds extracted by chemical software or algorithms~\citep{surface4}, have also been proposed~\citep{surface1,surface2,surface3}.  While biomolecular surfaces provide additional complements, they are not as natural and fundamental as electron clouds andd hardly reveal the underlying interactions between biomolecules.
 
\textbf{Learning interaction.}  Interaction between biomolecules ~\citep{du2016insights} are widely explored in the machine learning community~\citep{luo2023rotamer, somnath2021multi, wang2022learning}, which are ususlly based on SE(3)-equivariance~\citep{liao2023equiformer, equiformerv2, gotenet}. For molecules as ligands, they are usually represented as graphs at full-atom level~\citep{ hoogeboom2022equivariant, xu2022geodiff, lindiffbp} or fragments~\citep{geng2023novo, jin2018junction,lind3fg}. 
Besides, hierarchical ones are also proposed, achieving the state-of-the-art performance~\citep{get,hegnn}. These models are established with mostly supervised training. For unsupervised ones~\citep{dsmbind1, dsmbind2} including pretraining~\citep{promptddg, surface2}, they aim to enable the model to perceive structural flexibility. However, the electronic structures~\citep{ecloudgen, adams2024shepherddiffusingshapeelectrostatics} that govern the binding modes have largely been unexplored in the prediction of binding affinity.

\textbf{Relation to neural operators.} The proposed task of reconstruction of electron cloud density is similar to the motivation of neural operators~\citep{fno,neuraloperator, neuraloperator2}.  Firstly, neural operators are well-suited for processing signals with varying resolutions. Under the irregularly spatially sampled setting, our \textit{random selection as input augmentation} in Sec.~\ref{sec:encoder} aligns closely with the strengths of graph kernel neural operators~\citep{gkno}. Furthermore, the transformer with an attention mechanism is inherently a type of neural operator~\citep{wright2021transformersdeepinfinitedimensionalnonmercer}, which employs non-Mercer kernel functions to adaptively learn message-passing patterns through iterative updates~\citep{vit,afno}. 
Additionally, in the decoding process from patch centers to electron cloud signals, the coarse-to-fine message-passing resembles the multi-resolution matrix factorization employed in multipole graph neural operators~\citep{mgno}. The effectiveness of such network architectures and training augmentations in neural operators has been extensively validated~\citep{nonequifno, forecastnet}, and serves as one of the key inspirations for our design.

\vspace{-0.5em}
\section{Experiments} \label{sec:exp}

\begin{figure*}[ht]\vspace{-1em}
\begin{minipage}{0.7\linewidth} \centering
\captionof{table}{Results of relative binding affinity prediction on MISATO. \textbf{Bold} metrics are the best, and the \underline{underlined} values are the best baseline except ECBind.} \vspace{-0.5em}
\label{tab:relativebinding}
\resizebox{1\linewidth}{!}{\begin{tabular}{cccccccc}
\toprule
\multicolumn{2}{c}{\multirow{2}{*}{\diagbox{Methods}{Metrics}}}  & \multicolumn{4}{c}{Overall}                                       & \multicolumn{2}{c}{Per-structure} \\
\cmidrule(lr){3-6} \cmidrule(lr){7-8}
\multicolumn{2}{l}{}                          & Pearson($\uparrow$)        & Spearman($\uparrow$)        & RMSE($\downarrow$)            & AUROC($\uparrow$)           & Pearson($\uparrow$)          & Spearman($\uparrow$)         \\
\midrule
Benchmark                       & GCN         & 0.502\scriptsize{$\pm$0.038}          & 0.421\scriptsize{$\pm$0.043}          & 4.388\scriptsize{$\pm$0.127}          & 0.741\scriptsize{$\pm$0.020}          & 0.163\scriptsize{$\pm$0.015}           & 0.182\scriptsize{$\pm$0.018}           \\
\midrule
\multirow{9}{*}{Fragment}       & EGNN        & 0.579\scriptsize{$\pm$0.037}          & 0.535\scriptsize{$\pm$0.012}          & 4.572\scriptsize{$\pm$0.124}          & 0.759\scriptsize{$\pm$0.019}          & 0.297\scriptsize{$\pm$0.029}           & 0.244\scriptsize{$\pm$0.038}           \\
                                & SchNet      & 0.590\scriptsize{$\pm$0.024}          & 0.474\scriptsize{$\pm$0.012}          & 4.566\scriptsize{$\pm$0.052}          & 0.677\scriptsize{$\pm$0.007}          & 0.322\scriptsize{$\pm$0.036}           & 0.307\scriptsize{$\pm$0.030}           \\
                                & TorchMD     & 0.522\scriptsize{$\pm$0.038}          & 0.392\scriptsize{$\pm$0.052}          & 4.465\scriptsize{$\pm$0.112}          & 0.661\scriptsize{$\pm$0.034}          & 0.273\scriptsize{$\pm$0.047}           & 0.242\scriptsize{$\pm$0.053}           \\
                                & LEFTNet     & 0.639\scriptsize{$\pm$0.043}          & 0.497\scriptsize{$\pm$0.059}          & 4.385\scriptsize{$\pm$0.091}          & 0.717\scriptsize{$\pm$0.019}          & 0.284\scriptsize{$\pm$0.024}           & 0.267\scriptsize{$\pm$0.021}           \\
                                & GemNet      & 0.620\scriptsize{$\pm$0.035}          & 0.385\scriptsize{$\pm$0.089}          & 4.285\scriptsize{$\pm$0.040}          & 0.624\scriptsize{$\pm$0.056}          & 0.308\scriptsize{$\pm$0.039}           & 0.292\scriptsize{$\pm$0.011}           \\
                                & DimeNet     & 0.446\scriptsize{$\pm$0.063}          & 0.428\scriptsize{$\pm$0.035}          & 4.762\scriptsize{$\pm$0.138}          & 0.695\scriptsize{$\pm$0.033}          & 0.024\scriptsize{$\pm$0.083}           & 0.051\scriptsize{$\pm$0.041}           \\
                                & Equiformer  & \underline{0.666}\scriptsize{$\pm$0.020}          & 0.499\scriptsize{$\pm$0.045}          & 4.371\scriptsize{$\pm$0.063}          & 0.698\scriptsize{$\pm$0.051}          & 0.339\scriptsize{$\pm$0.011}           & 0.302\scriptsize{$\pm$0.044}           \\
                                & MACE        & 0.592\scriptsize{$\pm$0.044}          & 0.427\scriptsize{$\pm$0.035}          & 4.740\scriptsize{$\pm$0.055}          & 0.699\scriptsize{$\pm$0.055}          & 0.316\scriptsize{$\pm$0.052}           & 0.262\scriptsize{$\pm$0.018}           \\
                                & GET         & 0.651\scriptsize{$\pm$0.030}          & 0.473\scriptsize{$\pm$0.005}          & 4.494\scriptsize{$\pm$0.056}          & 0.673\scriptsize{$\pm$0.009}          & 0.366\scriptsize{$\pm$0.013}           & 0.359\scriptsize{$\pm$0.009}           \\
                                & HEGNN      & 0.555\scriptsize{$\pm$0.083}          & 0.497\scriptsize{$\pm$0.081}          & 4.467\scriptsize{$\pm$0.112}          & 0.697\scriptsize{$\pm$0.054}          & 0.262\scriptsize{$\pm$0.011}           & 0.235\scriptsize{$\pm$0.002}           \\
\midrule
\multirow{8}{*}{Atom}           & EGNN        & 0.378\scriptsize{$\pm$0.099}          & 0.321\scriptsize{$\pm$0.048}          & 4.938\scriptsize{$\pm$0.068}          & 0.548\scriptsize{$\pm$0.015}          & 0.248\scriptsize{$\pm$0.005}           & 0.242\scriptsize{$\pm$0.042}           \\
                                & SchNet      & 0.396\scriptsize{$\pm$0.071}          & 0.344\scriptsize{$\pm$0.073}          & 4.901\scriptsize{$\pm$0.130}          & 0.597\scriptsize{$\pm$0.056}          & 0.222\scriptsize{$\pm$0.028}           & 0.233\scriptsize{$\pm$0.023}           \\
                                & TorchMD     & 0.427\scriptsize{$\pm$0.131}          & 0.351\scriptsize{$\pm$0.115}          & 4.829\scriptsize{$\pm$0.110}          & 0.703\scriptsize{$\pm$0.067}          & 0.155\scriptsize{$\pm$0.031}           & 0.145\scriptsize{$\pm$0.035}           \\
                                & LEFTNet     & 0.364\scriptsize{$\pm$0.098}          & 0.270\scriptsize{$\pm$0.037}          & 4.881\scriptsize{$\pm$0.050}          & 0.517\scriptsize{$\pm$0.019}          & 0.196\scriptsize{$\pm$0.025}           & 0.173\scriptsize{$\pm$0.057}           \\
                                & GemNet     & 0.578\scriptsize{$\pm$0.048}          & 0.376\scriptsize{$\pm$0.072}          & 4.624\scriptsize{$\pm$0.090}          & 0.660\scriptsize{$\pm$0.050}          & 0.375\scriptsize{$\pm$0.032}           & 0.348\scriptsize{$\pm$0.056}           \\
                                & DimeNet     & 0.393\scriptsize{$\pm$0.147}          & 0.336\scriptsize{$\pm$0.217}          & 5.040\scriptsize{$\pm$0.014}          & 0.677\scriptsize{$\pm$0.045}          & 0.071\scriptsize{$\pm$0.033}           & 0.100\scriptsize{$\pm$0.062}           \\
                                & Equiformer  & OOM            & OOM            & OOM            & OOM            & OOM             & OOM             \\
                                & MACE     & 0.452\scriptsize{$\pm$0.172}          & 0.359\scriptsize{$\pm$0.162}          & 4.469\scriptsize{$\pm$0.351}          & 0.641\scriptsize{$\pm$0.099}          & 0.237\scriptsize{$\pm$0.010}           & 0.194\scriptsize{$\pm$0.065}           \\
                                & GET         & 0.443\scriptsize{$\pm$0.079}          & 0.307\scriptsize{$\pm$0.089}          & 4.913\scriptsize{$\pm$0.080}          & 0.582\scriptsize{$\pm$0.007}          &        0.294\scriptsize{$\pm$0.048}           & 0.267\scriptsize{$\pm$0.030}           \\
                                % 0.226\scriptsize{$\pm$0.000}           & 0.230\scriptsize{$\pm$0.000}           \\
                                & HEGNN      & 0.545\scriptsize{$\pm$0.089}          & 0.486\scriptsize{$\pm$0.069}          & 4.501\scriptsize{$\pm$0.077}          & 0.705\scriptsize{$\pm$0.049}          & 0.267\scriptsize{$\pm$0.027}           & 0.271\scriptsize{$\pm$0.039}           \\
\midrule
\multirow{8}{*}{Hierarchical}   & EGNN        & 0.643\scriptsize{$\pm$0.003}          & 0.531\scriptsize{$\pm$0.044}          & 4.265\scriptsize{$\pm$0.130}          & 0.723\scriptsize{$\pm$0.045}          & 0.333\scriptsize{$\pm$0.056}           & 0.355\scriptsize{$\pm$0.041}           \\
                                & SchNet      & 0.601\scriptsize{$\pm$0.098}          & 0.520\scriptsize{$\pm$0.033}          & 4.328\scriptsize{$\pm$0.137}          & 0.709\scriptsize{$\pm$0.013}          & 0.350\scriptsize{$\pm$0.048}           & 0.336\scriptsize{$\pm$0.065}           \\
                                & TorchMD     & 0.597\scriptsize{$\pm$0.029}          & 0.420\scriptsize{$\pm$0.034}          & 4.253\scriptsize{$\pm$0.131}          & 0.670\scriptsize{$\pm$0.031}          & 0.329\scriptsize{$\pm$0.015}           & 0.316\scriptsize{$\pm$0.031}           \\
                                & LEFTNet     & 0.611\scriptsize{$\pm$0.026}          & 0.476\scriptsize{$\pm$0.059}          & 4.230\scriptsize{$\pm$0.051}         & 0.718\scriptsize{$\pm$0.058}          & 0.350\scriptsize{$\pm$0.015}           & 0.374\scriptsize{$\pm$0.054}          \\
                                & GemNet     & 0.655\scriptsize{$\pm$0.067}          & 0.547\scriptsize{$\pm$0.080}          & 4.135\scriptsize{$\pm$0.113}          & 0.769\scriptsize{$\pm$0.068}          & 0.248\scriptsize{$\pm$0.108}           & 0.244\scriptsize{$\pm$0.122}           \\
                                & DimeNet     & 0.599\scriptsize{$\pm$0.020}          & 0.509\scriptsize{$\pm$0.059}          & 4.607\scriptsize{$\pm$0.049}          & 0.699\scriptsize{$\pm$0.026}          & 0.346\scriptsize{$\pm$0.016}           & 0.293\scriptsize{$\pm$0.098}           \\
                                & Equiformer  & OOM            & OOM            & OOM            & OOM            & OOM             & OOM             \\
                                & MACE     & 0.608\scriptsize{$\pm$0.097}          & \underline{0.549}\scriptsize{$\pm$0.048}          & \underline{4.035}\scriptsize{$\pm$0.464}          & \underline{0.783}\scriptsize{$\pm$0.052}          & 0.240\scriptsize{$\pm$0.114}           & 0.206\scriptsize{$\pm$0.084}           \\
                                & GET         & 0.646\scriptsize{$\pm$0.016}          & 0.472\scriptsize{$\pm$0.048}          & 4.423\scriptsize{$\pm$0.063}          & 0.657\scriptsize{$\pm$0.028}          & \underline{0.405}\scriptsize{$\pm$0.019}           & \underline{0.385}\scriptsize{$\pm$0.056}           \\
                                & HEGNN      & 0.590\scriptsize{$\pm$0.028}          & 0.535\scriptsize{$\pm$0.029}          & 4.333\scriptsize{$\pm$0.253}          & 0.762\scriptsize{$\pm$0.034}          & 0.313\scriptsize{$\pm$0.013}           & 0.291\scriptsize{$\pm$0.035}           \\
\midrule
\multirow{3}{*}{Ours} & ECBind-nptn  & 0.621\scriptsize{$\pm$0.026}          & 0.554\scriptsize{$\pm$0.014}          & 3.936\scriptsize{$\pm$0.025}          & 0.759\scriptsize{$\pm$0.014}          & 0.283\scriptsize{$\pm$0.016}           & 0.258\scriptsize{$\pm$0.028}           \\
                                & ECBind-stdt & \textbf{0.697}\scriptsize{$\pm$0.019}  & 0.646\scriptsize{$\pm$0.008}          & 3.907\scriptsize{$\pm$0.122}          & 0.824\scriptsize{$\pm$0.021}          & 0.407\scriptsize{$\pm$0.027}           & 0.420\scriptsize{$\pm$0.025}           \\
                                & ECBind-ptn      & 0.692\scriptsize{$\pm$0.013}          & \textbf{0.665}\scriptsize{$\pm$0.014} & \textbf{3.892}\scriptsize{$\pm$0.044} & \textbf{0.843}\scriptsize{$\pm$0.033} & \textbf{0.431}\scriptsize{$\pm$0.022}  & \textbf{0.445}\scriptsize{$\pm$0.034}  \\
\bottomrule
\end{tabular}}
\end{minipage}\vspace{1em}
\begin{minipage}{0.29\linewidth}
    \centering\vspace{0.5em}
        
            \includegraphics[height=0.45\linewidth, trim=10 40 40 10, clip]{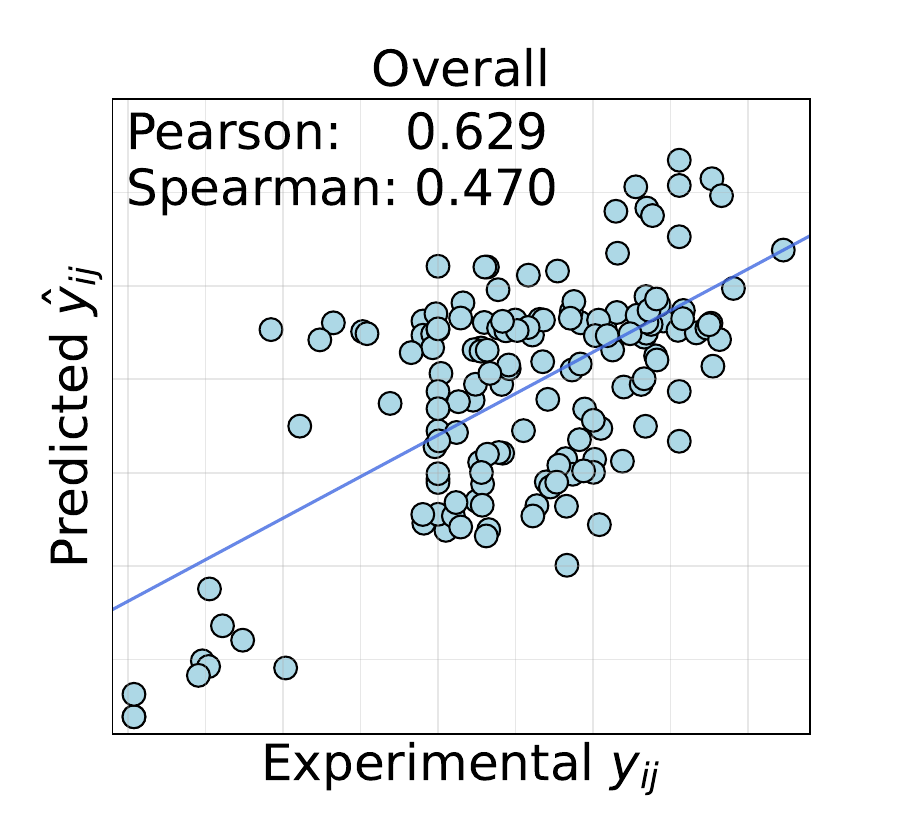}
            \label{fig:oaget}
            \includegraphics[height=0.45\linewidth, trim=49 40 40 10, clip]{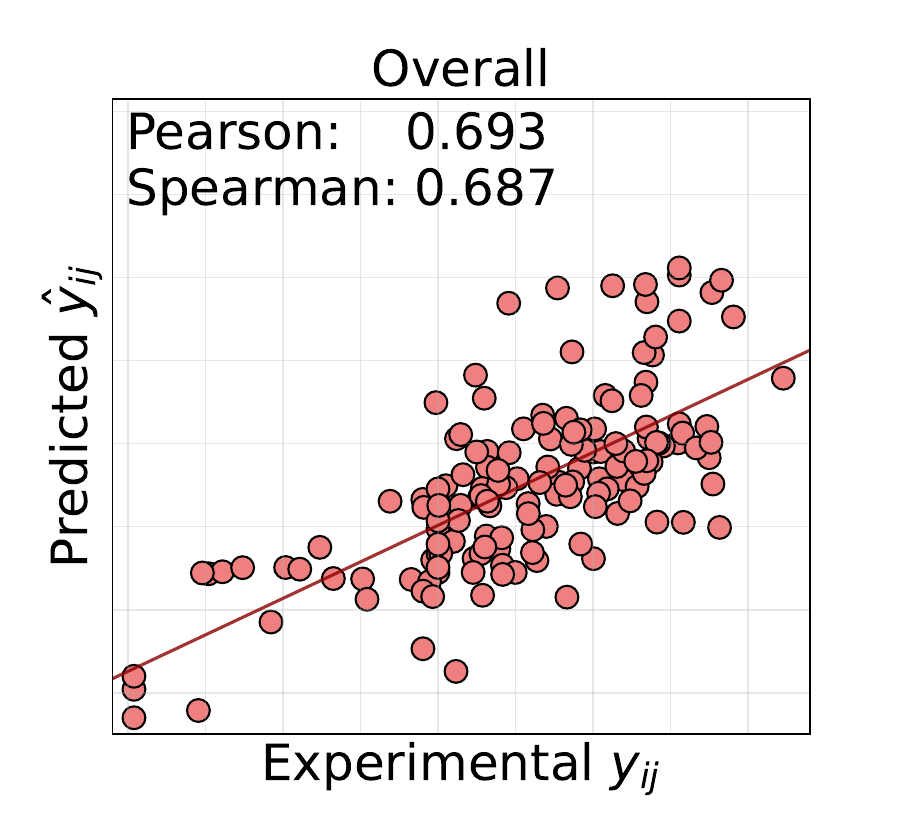}
            \label{fig:oaecbind}\\\vspace{-1.1em}
            \includegraphics[height=0.415\linewidth, trim=10 35 40 10, clip]{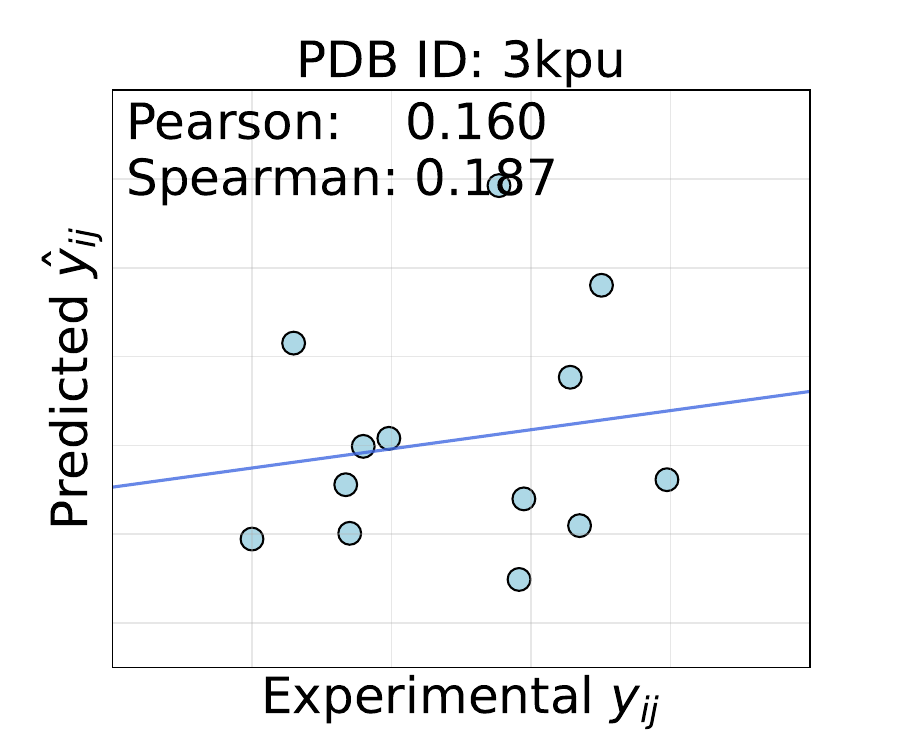}
            \label{fig:3kpuget}
            \includegraphics[height=0.415\linewidth, trim=49 35 40 10, clip]{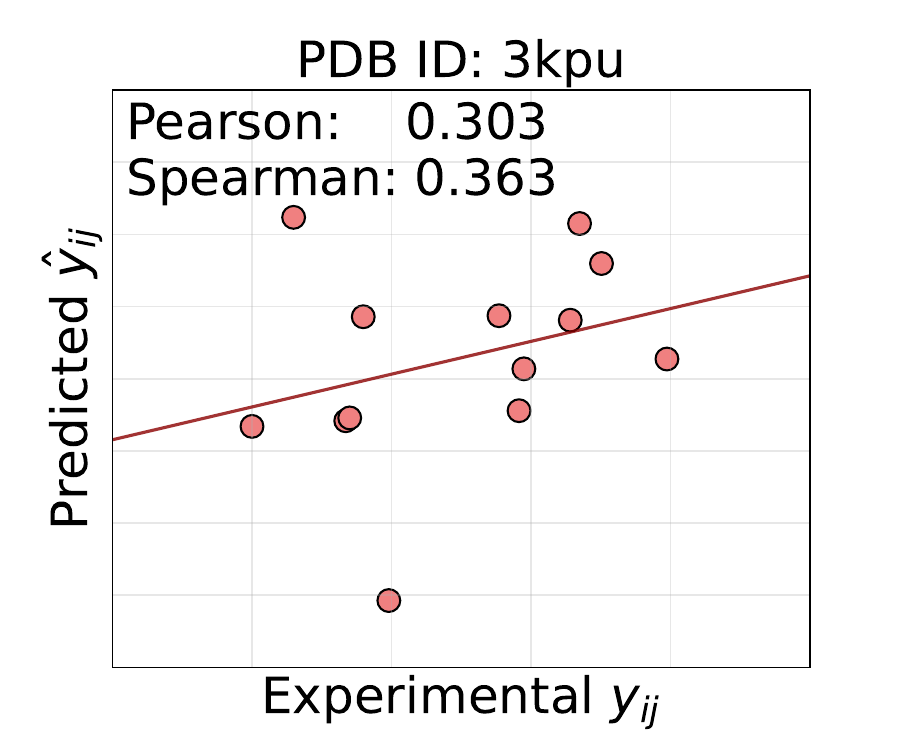}
            \label{fig:3kpuecbind} \\\vspace{-1.1em}
            \includegraphics[height=0.415\linewidth, trim=10 35 40 10, clip]{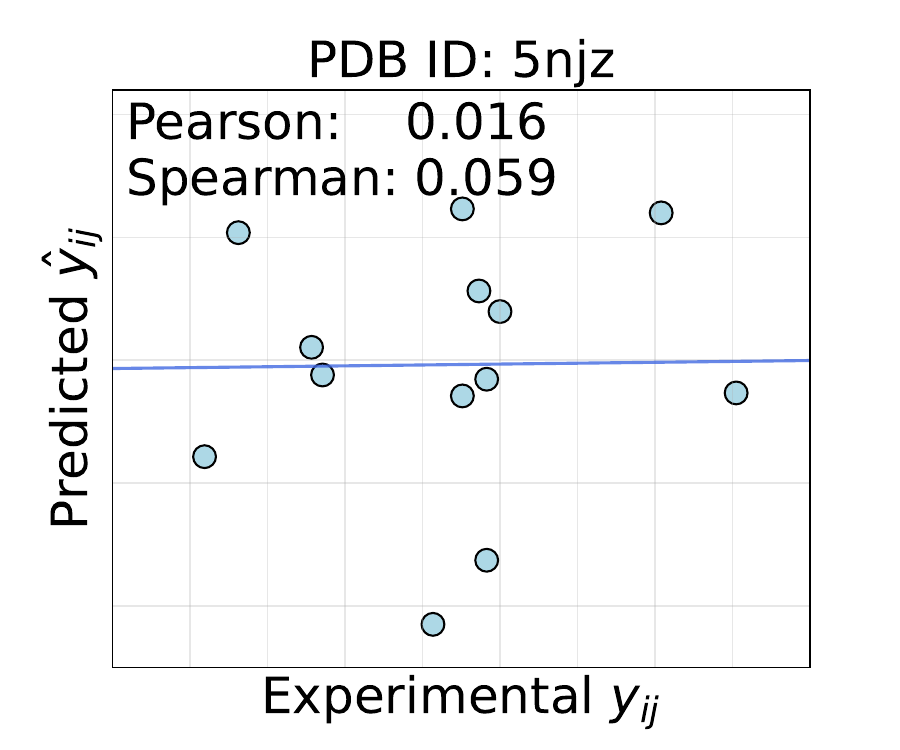}
            \label{fig:5njzget}
            \includegraphics[height=0.415\linewidth, trim=49 35 40 10, clip]{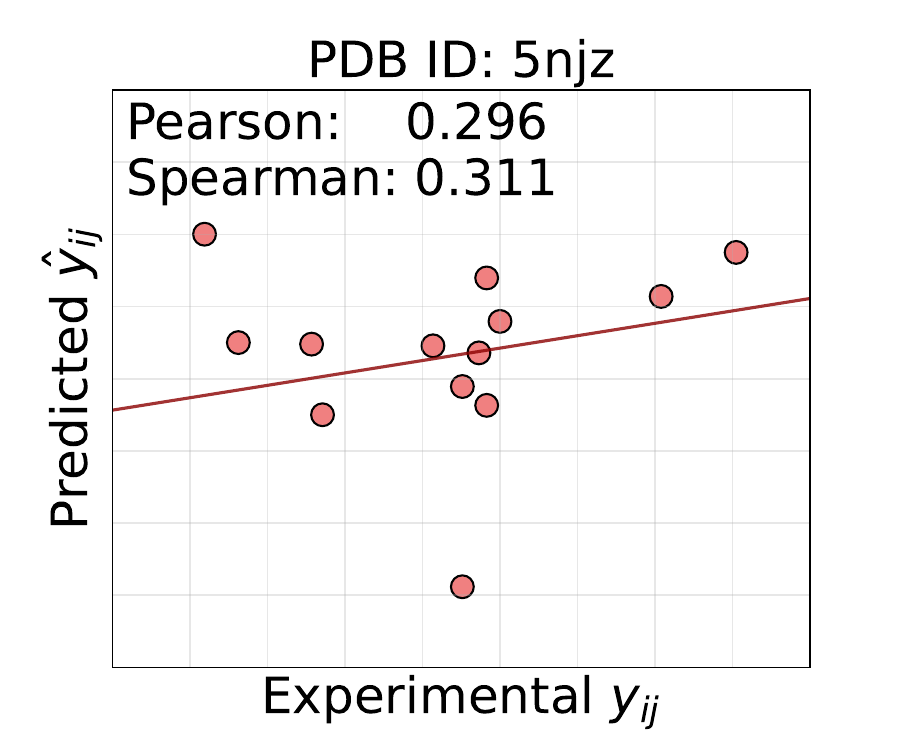}
            \\
            \includegraphics[height=0.411\linewidth, trim=10 35 40 10, clip]{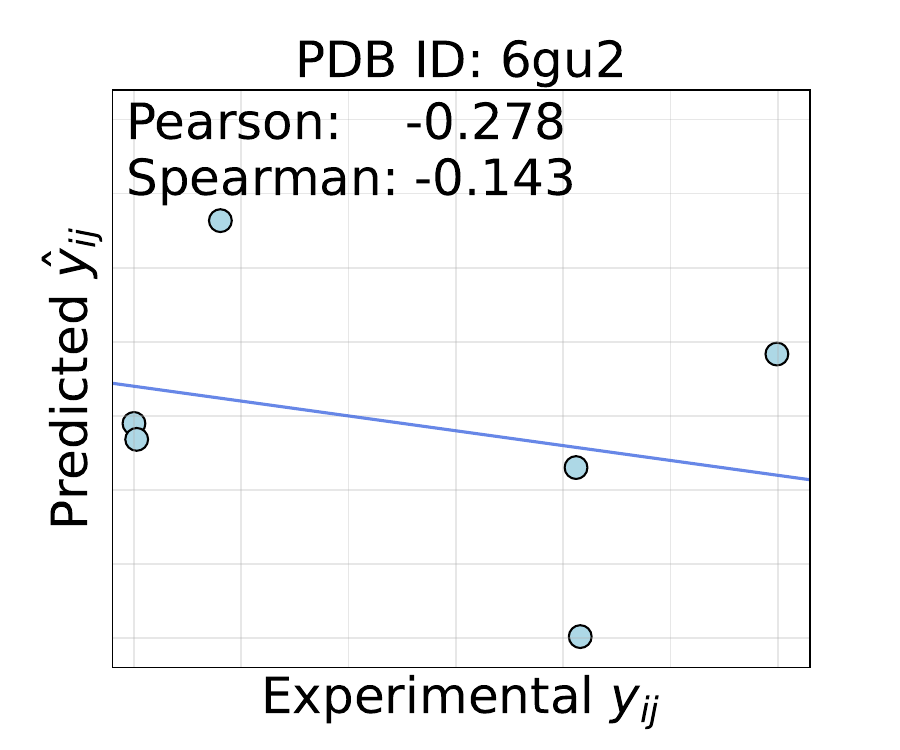}
            \label{fig:6gu2get}
            \includegraphics[height=0.411\linewidth, trim=49 35 40 10, clip]{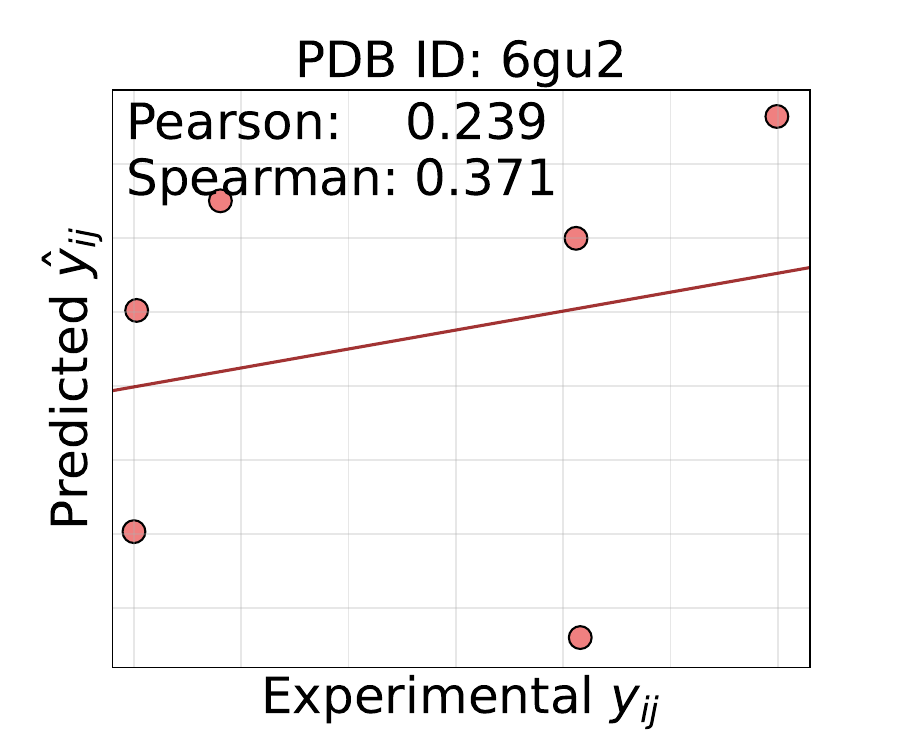}
            \label{fig:6gu2ecbind}
            \includegraphics[height=0.44\linewidth, trim=10 12 42.5 10, clip]{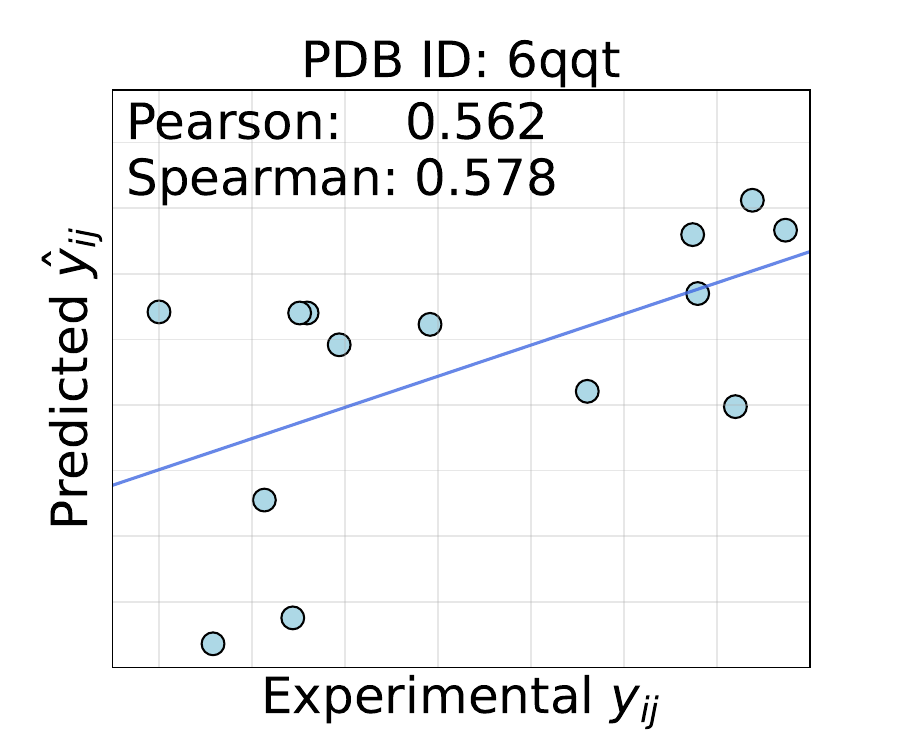}
            \label{fig:6qqtecbind}
            \includegraphics[height=0.44\linewidth, trim=49 12 42.5 10, clip]{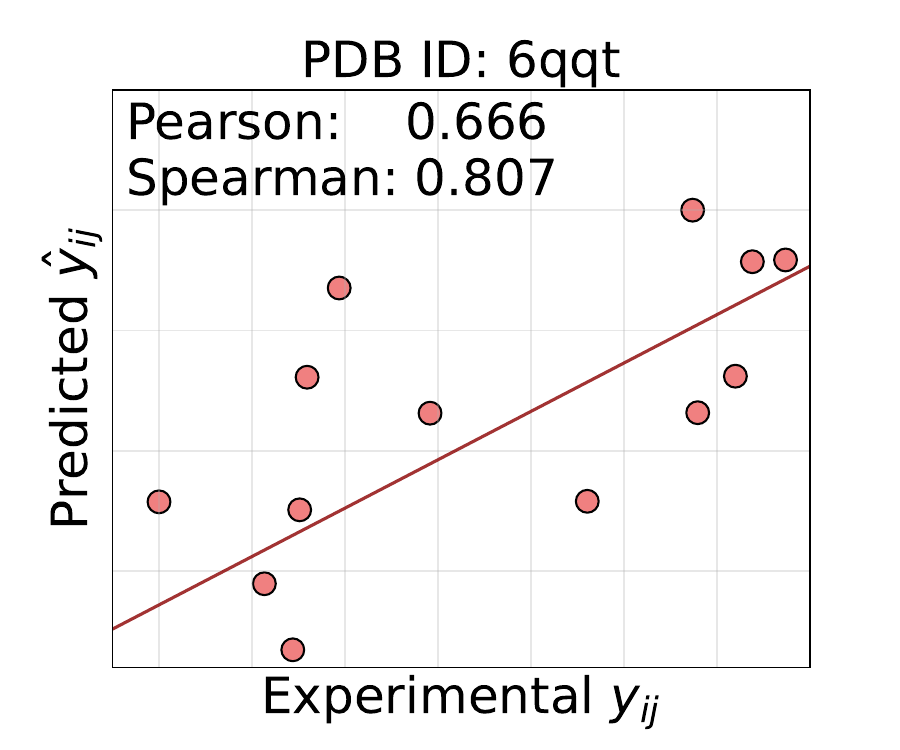}
            \label{fig:6qqtget}\vspace{-1.5em}
            \caption{Visualization: GET~(blue) and ECBind~(red) on per-structure metrics for `hard' protein samples.}
        \label{fig:deltabind}
\end{minipage}
\vspace{-2.5em}
\end{figure*}
\subsection{Setup}
\vspace{-0.5em}
\textbf{Tasks and Datasets.} We evaluate the effectiveness of the proposed ECBind on three tasks and their corresponding datasets. First, we follow the protocol from \textbf{Atom3D}~\citep{atom3d}, which includes all protein amino acids within a distance of $\leq 6\ \text{\AA}$ from the atoms in the molecule to form the pocket. We account for the positions of hydrogen atoms, which is crucial for accurate electron cloud computation. Specifically, we use \textbf{Reduce}~\citep{reduce} to add missing hydrogen atoms in the proteins' amino acids and compute their coordinates, while \textbf{MMFF}~\citep{mmff} is employed to estimate the coordinates of missing hydrogens in the molecule. After obtaining the refined binding conformations, we use \textbf{XTB}~\citep{xtb} to compute the electron cloud density. Structures that fail to converge during computation are treated as low-quality data and excluded from the datasets. This leads to the listed tasks and datasets: 
\begin{itemize}
    \item 
    [(i)] \textbf{Relative binding affinity with MISATO}~\citep{misato}: It refines PDBBind~\citep{pdbbind} using molecular dynamics simulations of 19,443 experimental protein-ligand complexes. However, we exclude the quantum mechanical properties it provides, as they are challenging to obtain in real-world scenarios. After processes, including the removal of data without labels and filtering based on convergence during electron cloud computation, the final training, validation, and test sets are split according to the benchmark\footnote{{https://github.com/kierandidi/misato-affinity}}, resulting in 9,964, 1,074, and 156 pairs.
    
    \item [(ii)] \textbf{Absolute binding affinity with LBA}~\citep{lba}: Using a sequence identity split of 30\%, the included pairs are processed similarly to (i), resulting in 3,067, 346, and 426 instances for training, validation, and test.  

    \item [(iii)] \textbf{Ligand specificity with LEP}~\citep{glide}: With split by functional proteins, the dataset yields 536, 202, and 162 pairs for training, validation, and test sets.
\end{itemize}
\begin{figure*} [ht] \vspace{-0.5em}
\begin{minipage}{0.65\linewidth}
\centering
\captionof{table}{Results of metrics on LBA and LEP. Since they are absolute binding prediction tasks, `per-structure' metrics are not feasible. \textbf{Bold} metrics are the best, and the \underline{underlined} values are the best baseline except ECBind.} \vspace{-0.5em}
\label{tab:lba_lep}
\resizebox{1\linewidth}{!}{\begin{tabular}{cccccccc}
\toprule
\multicolumn{2}{c}{\multirow{2}{*}{\diagbox{Methods}{Metrics}}}  & \multicolumn{3}{c}{LBA}                                       & \multicolumn{2}{c}{LEP} \\
\cmidrule(lr){3-5} \cmidrule(lr){6-7}
\multicolumn{2}{l}{}                          & Pearson($\uparrow$)        & Spearman($\uparrow$)        & RMSE($\downarrow$)            & AUROC($\uparrow$)           & AUPRC($\uparrow$)\\
\midrule
\multirow{7}{*}{Block} 
& SchNet       & 0.565\scriptsize{$\pm$0.006} & 0.549\scriptsize{$\pm$0.007} & 1.406\scriptsize{$\pm$0.020} & 0.732\scriptsize{$\pm$0.022} & 0.718\scriptsize{$\pm$0.031} \\
& DimeNet++    & 0.576\scriptsize{$\pm$0.016} & 0.569\scriptsize{$\pm$0.016} & 1.391\scriptsize{$\pm$0.020} & 0.669\scriptsize{$\pm$0.014} & 0.609\scriptsize{$\pm$0.036} \\
& EGNN         & 0.566\scriptsize{$\pm$0.010} & 0.548\scriptsize{$\pm$0.012} & 1.409\scriptsize{$\pm$0.015} & 0.746\scriptsize{$\pm$0.017} & 0.755\scriptsize{$\pm$0.031} \\
& ET           & 0.599\scriptsize{$\pm$0.017} & 0.584\scriptsize{$\pm$0.025} & 1.367\scriptsize{$\pm$0.037} & 0.744\scriptsize{$\pm$0.034} & 0.721\scriptsize{$\pm$0.052} \\
& GemNet       & 0.569\scriptsize{$\pm$0.027} & 0.553\scriptsize{$\pm$0.026} & 1.393\scriptsize{$\pm$0.036} & 0.590\scriptsize{$\pm$0.062} & 0.541\scriptsize{$\pm$0.107} \\
& MACE         & 0.599\scriptsize{$\pm$0.010} & 0.580\scriptsize{$\pm$0.014} & 1.385\scriptsize{$\pm$0.006} & 0.617\scriptsize{$\pm$0.022} & 0.549\scriptsize{$\pm$0.007} \\
& Equiformer         & 0.604\scriptsize{$\pm$0.013} & 0.591\scriptsize{$\pm$0.012} & 1.350\scriptsize{$\pm$0.019} & 0.000\scriptsize{$\pm$0.000} & 0.000\scriptsize{$\pm$0.000} \\
& LEFTNet      & 0.588\scriptsize{$\pm$0.011} & 0.576\scriptsize{$\pm$0.010} & 1.377\scriptsize{$\pm$0.013} & 0.738\scriptsize{$\pm$0.001} & 0.750\scriptsize{$\pm$0.006} \\
& GET     & 0.611\scriptsize{$\pm$0.005} & 0.610\scriptsize{$\pm$0.010} & 1.356\scriptsize{$\pm$0.004} & 0.766\scriptsize{$\pm$0.010} & 0.717\scriptsize{$\pm$0.033} \\
& HEGNN         & 0.589\scriptsize{$\pm$0.013} & 0.607\scriptsize{$\pm$0.011} & 1.422\scriptsize{$\pm$0.023} & 0.739\scriptsize{$\pm$0.003} & 0.710\scriptsize{$\pm$0.038} \\
\midrule
\multirow{7}{*}{Atom}
& SchNet       & 0.598\scriptsize{$\pm$0.011} & 0.592\scriptsize{$\pm$0.015} & 1.357\scriptsize{$\pm$0.017} & 0.712\scriptsize{$\pm$0.026} & 0.639\scriptsize{$\pm$0.033} \\
& DimeNet++       & 0.547\scriptsize{$\pm$0.015} & 0.536\scriptsize{$\pm$0.016} & 1.439\scriptsize{$\pm$0.036} & 0.589\scriptsize{$\pm$0.049} & 0.503\scriptsize{$\pm$0.020} \\
& EGNN         & 0.599\scriptsize{$\pm$0.002} & 0.587\scriptsize{$\pm$0.004} & 1.358\scriptsize{$\pm$0.000} & 0.711\scriptsize{$\pm$0.020} & 0.643\scriptsize{$\pm$0.041} \\
& ET           & 0.591\scriptsize{$\pm$0.007} & 0.583\scriptsize{$\pm$0.009} & 1.381\scriptsize{$\pm$0.013} & 0.677\scriptsize{$\pm$0.004} & 0.636\scriptsize{$\pm$0.054} \\
& GemNet           & 0.610\scriptsize{$\pm$0.004} & 0.580\scriptsize{$\pm$0.007} & 1.341\scriptsize{$\pm$0.009} &0.575\scriptsize{$\pm$0.012} & 0.482\scriptsize{$\pm$0.033} \\
& MACE         & 0.579\scriptsize{$\pm$0.009} & 0.563\scriptsize{$\pm$0.012} & 1.411\scriptsize{$\pm$0.029} & 0.614\scriptsize{$\pm$0.047} & 0.571\scriptsize{$\pm$0.046} \\
& Equiformer           & OOM & OOM & OOM & OOM & OOM \\
& LEFTNet      & 0.610\scriptsize{$\pm$0.004} & 0.598\scriptsize{$\pm$0.003} & 1.343\scriptsize{$\pm$0.004} & 0.740\scriptsize{$\pm$0.024} & 0.698\scriptsize{$\pm$0.026} \\
& GET     & 0.613\scriptsize{$\pm$0.010} & \underline{0.618}\scriptsize{$\pm$0.009} & \underline{1.309}\scriptsize{$\pm$0.014} & \underline{0.769}\scriptsize{$\pm$0.027} & 0.736\scriptsize{$\pm$0.034} \\
& HEGNN         & 0.591\scriptsize{$\pm$0.007} & 0.596\scriptsize{$\pm$0.006} & 1.521\scriptsize{$\pm$0.017} & 0.725\scriptsize{$\pm$0.056} & 0.638\scriptsize{$\pm$0.099} \\
\midrule
\multirow{7}{*}{Hierarchical}
& SchNet       & 0.590\scriptsize{$\pm$0.017} & 0.571\scriptsize{$\pm$0.028} & 1.370\scriptsize{$\pm$0.028} & 0.736\scriptsize{$\pm$0.020} & 0.731\scriptsize{$\pm$0.048} \\
& DimeNet++       & 0.582\scriptsize{$\pm$0.009} & 0.574\scriptsize{$\pm$0.007} & 1.388\scriptsize{$\pm$0.010} & 0.579\scriptsize{$\pm$0.118} & 0.517\scriptsize{$\pm$0.100} \\
& EGNN         & 0.586\scriptsize{$\pm$0.004} & 0.568\scriptsize{$\pm$0.004} & 1.380\scriptsize{$\pm$0.015} & 0.724\scriptsize{$\pm$0.027} & 0.720\scriptsize{$\pm$0.056} \\
& ET           & 0.580\scriptsize{$\pm$0.008} & 0.564\scriptsize{$\pm$0.004} & 1.383\scriptsize{$\pm$0.009} & 0.717\scriptsize{$\pm$0.033} & 0.724\scriptsize{$\pm$0.055} \\
& GemNet           & 0.535\scriptsize{$\pm$0.041} & 0.535\scriptsize{$\pm$0.016} & 1.455\scriptsize{$\pm$0.081} &0.546\scriptsize{$\pm$0.023} & 0.519\scriptsize{$\pm$0.070} \\
& MACE         & 0.612\scriptsize{$\pm$0.010} & 0.592\scriptsize{$\pm$0.010} & 1.372\scriptsize{$\pm$0.021} & 0.530\scriptsize{$\pm$0.032} & 0.456\scriptsize{$\pm$0.038} \\
& Equiformer           & OOM & OOM & OOM & OOM & OOM \\
& LEFTNet      & 0.592\scriptsize{$\pm$0.014} & 0.580\scriptsize{$\pm$0.011} & 1.366\scriptsize{$\pm$0.016} & 0.719\scriptsize{$\pm$0.020} & 0.708\scriptsize{$\pm$0.012} \\
& GET     &\underline{0.614}\scriptsize{$\pm$0.005} & 0.612\scriptsize{$\pm$0.013} & \underline{1.309}\scriptsize{$\pm$0.011} & 0.767\scriptsize{$\pm$0.017} & \underline{0.738}\scriptsize{$\pm$0.043} \\
& HEGNN         & 0.586\scriptsize{$\pm$0.005} & 0.592\scriptsize{$\pm$0.007} & 1.486\scriptsize{$\pm$0.011} & 0.703\scriptsize{$\pm$0.028} & 0.688\scriptsize{$\pm$0.030} \\
\midrule
\multirow{3}{*}{Ours} 
& ECBind-nptn   & {0.613}\scriptsize{$\pm$0.008} & {0.606}\scriptsize{$\pm$0.008} & {1.359}\scriptsize{$\pm$0.027} & {0.743}\scriptsize{$\pm$0.007} & {0.752}\scriptsize{$\pm$0.005} \\
& ECBind-stdt   & {0.614}\scriptsize{$\pm$0.004} & {0.615}\scriptsize{$\pm$0.003} & {1.314}\scriptsize{$\pm$0.005} & {0.768}\scriptsize{$\pm$0.006} & \textbf{0.777}\scriptsize{$\pm$0.000} \\
& ECBind-ptn   & \textbf{0.626}\scriptsize{$\pm$0.006} & \textbf{0.622}\scriptsize{$\pm$0.004} & \textbf{1.298}\scriptsize{$\pm$0.003} & \textbf{0.775}\scriptsize{$\pm$0.008} & {0.763}\scriptsize{$\pm$0.004} \\
\bottomrule
\end{tabular}}
\end{minipage}
\begin{minipage}{0.35\linewidth}

\refstepcounter{figure} \label{fig:tokenizertrain}
\label{fig:tokenizertrain}
\centering
         \subcaptionbox{
            Electron cloud tokenzier loss \label{fig:lossec} \vspace{-1.3em}
        }{\vspace{-0.7em}\hspace{-0.3em}
        \includegraphics[height=0.59\linewidth, trim=10 00 40 30, clip]{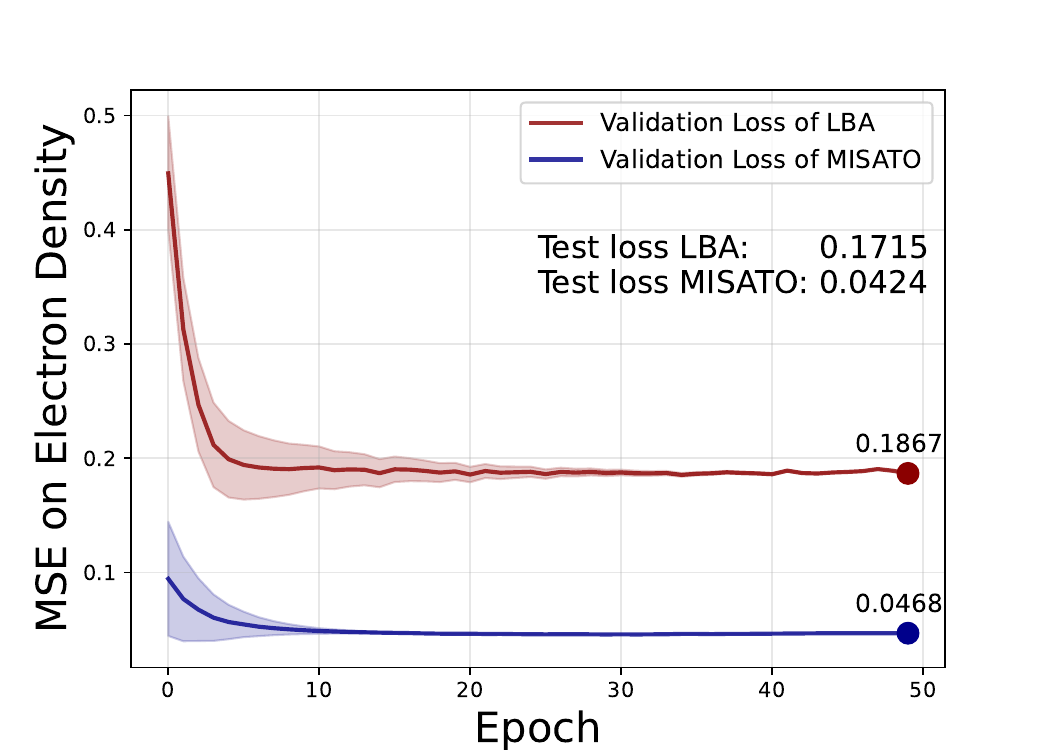}}
        
        \subcaptionbox{ 
            Full atom tokenizer loss \label{fig:lossfa}
        }{
        \vspace{-0.7em} \hspace{0.85em}
        \includegraphics[height=0.59\linewidth, trim=20 00 00 30, clip]{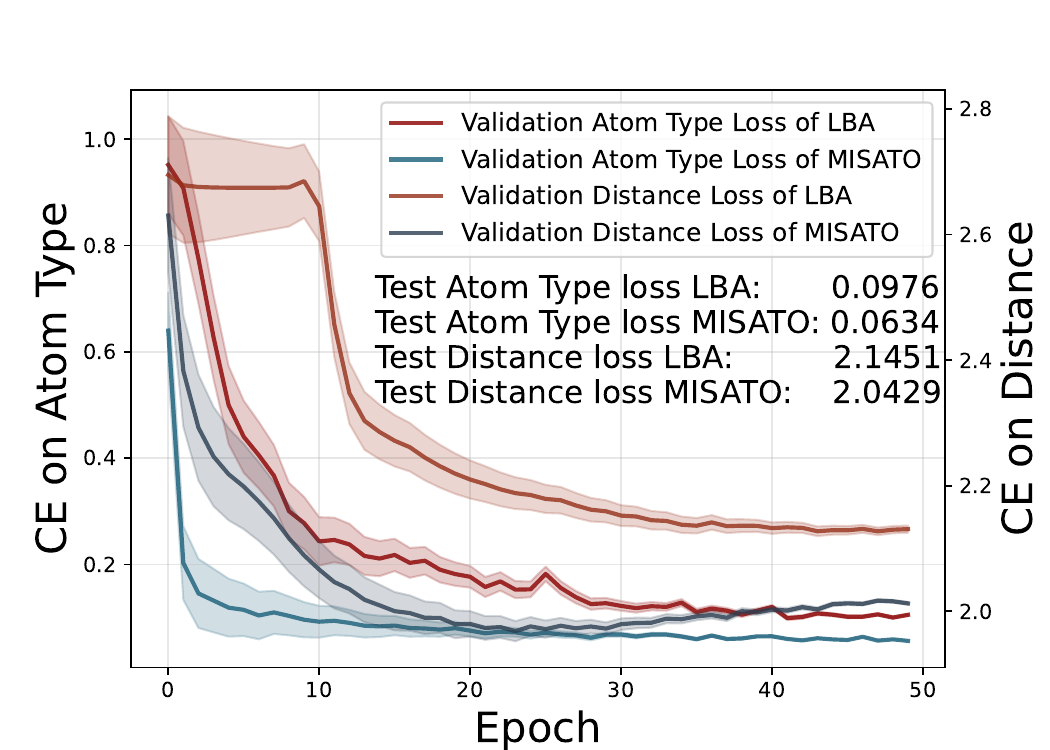}
        }
        \subcaptionbox{
            Overall correlations \label{fig:codeablation}
        }{\vspace{-0.7em} \hspace{-0.3em}
        \includegraphics[height=0.52\linewidth, trim=10 37 10 10, clip]{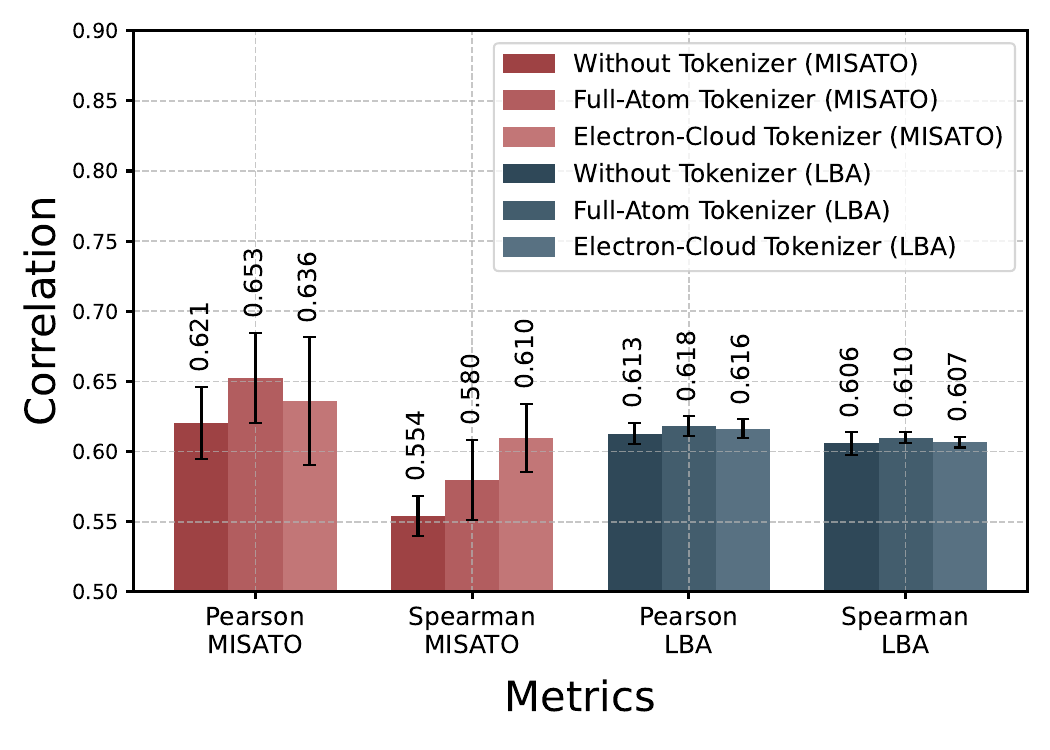}
        }\vspace{-0.5em}
\caption*{{Figure~\thefigure.} Losses in tokenizers when training and testing, and metrics of variants of ECBind with/without different tokenizers.}
\end{minipage}\vspace{-1.9em}
        
\end{figure*}

\textbf{Training and Baseline.} For performance comparison, we follow the protocols of \textbf{GET}~\citep{get}, benchmarking against GNN baselines including \textbf{SchNet}~\citep{schnet}, \textbf{DimNet++}~\citep{Klicpera2020DirectionalMP}, \textbf{GemNet}~\citep{Klicpera2021GemNetUD}, \textbf{TorchMD-Net}~\citep{tholke2022equivariant}, \textbf{LEFTNet}~\citep{du2023a}, \textbf{MACE}~\citep{batatia2022mace}, and \textbf{Equiformer}~\citep{liao2023equiformer}, and the latest proposed \textbf{HEGNN}~\citep{hegnn}. The descriptions of these models can be found in Appendix~\ref{app:methodintro}. Following the categorization from GET, we group the baselines into fragment-level, atom-level, and hierarchical message-passing methods.
In {ECBind}, we pretrain both atomic and electronic structures and finetune the lightweight transformer head on the same dataset for specific tasks, leading to \textbf{ECBind-ptn}. For other baseline models, they are directly trained on the training sets, with hyperparameters detailed in Appendix~\ref{app:training}. Additionally, we introduce two variations of ECBind: \textbf{ECBind-nptn}, a non-pretraining version that only leverages full-atom structure as inputs, to demonstrate the performance gains from pretraining and the additional gaining from electron cloud information, and \textbf{ECBind-stdt} as an electron-cloud-agnostic student version trained with knowledge distillation from ECBind-ptn.\vspace{0.4em}

\textbf{Metrics.} All reported metrics, presented in the format of \(\text{mean}\pm\text{std}\), are calculated based on three independent runs. Baselines that fail to process atomic graphs due to high complexity are marked with OOM (out-of-memory). To evaluate task (i) relative binding affinity, we consider \textbf{Pearson} and \textbf{Spearman} correlation coefficient, minimized \textbf{RMSE} (root mean squared error), and \textbf{AUROC} (area under the receiver operating characteristic). To calculate AUROC, affinities are classified based on the sign of $\hat{y}_{ij}$.  In practical applications, the correlation for one specific protein with different ligands is often of greater interest, so we evaluate average \textbf{per-structure Pearson} and \textbf{per-structure Spearman} correlation coefficient. To evaluate task (ii) absolute binding affinity, we consider Pearson and Spearman's correlation and RMSE. To evaluate task (iii) ligand specificity, we use AUROC, \textbf{AUPRC} (area under the precision-recall curve). \vspace{-0.5em}

\subsection{Results}
\textbf{MISATO relative binding affinity.} Table.~\ref{tab:relativebinding} gives the performance comparison of ECBind with different baseline models, and Figure.~\ref{fig:deltabind} gives the visualization in certain `hard' proteins with different molecules that models hardly give accurate predictions, from which we can conclude that 
\begin{itemize}
    \item 
[(i)] ECBind with pretrained tokens achieves the overall best according to the reported metrics, with 6.42\% and 15.58\% improvements on per-structure Pearson and Spearman correlation over the best baseline model, respectively. 
\item 
    [(ii)] The non-pretrained version shows comparable performance in overall metrics; However, it struggles to differentiate variations in affinity caused by different molecules binding to the same protein, as shown in its significantly lower per-structure Pearson and Spearman correlations. Additionally, the electron-cloud-agnostic student model exhibits a marginal performance drop compared to the teacher model, with  5.56\% and 5.61\% decreases in Pearson and Spearman correlation. 
    \item 
    [(iii)] Overall, models incorporating fragment-level graphs (Fragment\&Hierarchical) outperform those based solely on atom-level graphs. In comparison, ECBind integrates electron-cloud- and atom-level structures. This suggests a promising direction for our future work—incorporating amino acids and functional groups into ECBind to further enhance its capabilities. 

\end{itemize}
\textbf{LBA and LEP.}  Table.~\ref{tab:lba_lep} shows the results of binding affinity and specificity prediction on LBA and LEP and demonstrates the superiority of ECBind in the two tasks.  It is worth noting that while ECBind achieves the best performance in overall Pearson and Spearman correlations on the LBA, its improvement over the non-pretrained version and the previous state-of-the-art GET is not as significant on MISATO. This will be further discussed in the next section.

\subsection{Analysis}
\vspace{-0.5em}
\begin{figure}
    \centering{
    \includegraphics[width=0.49\textwidth, trim=20 90 120 120, clip]{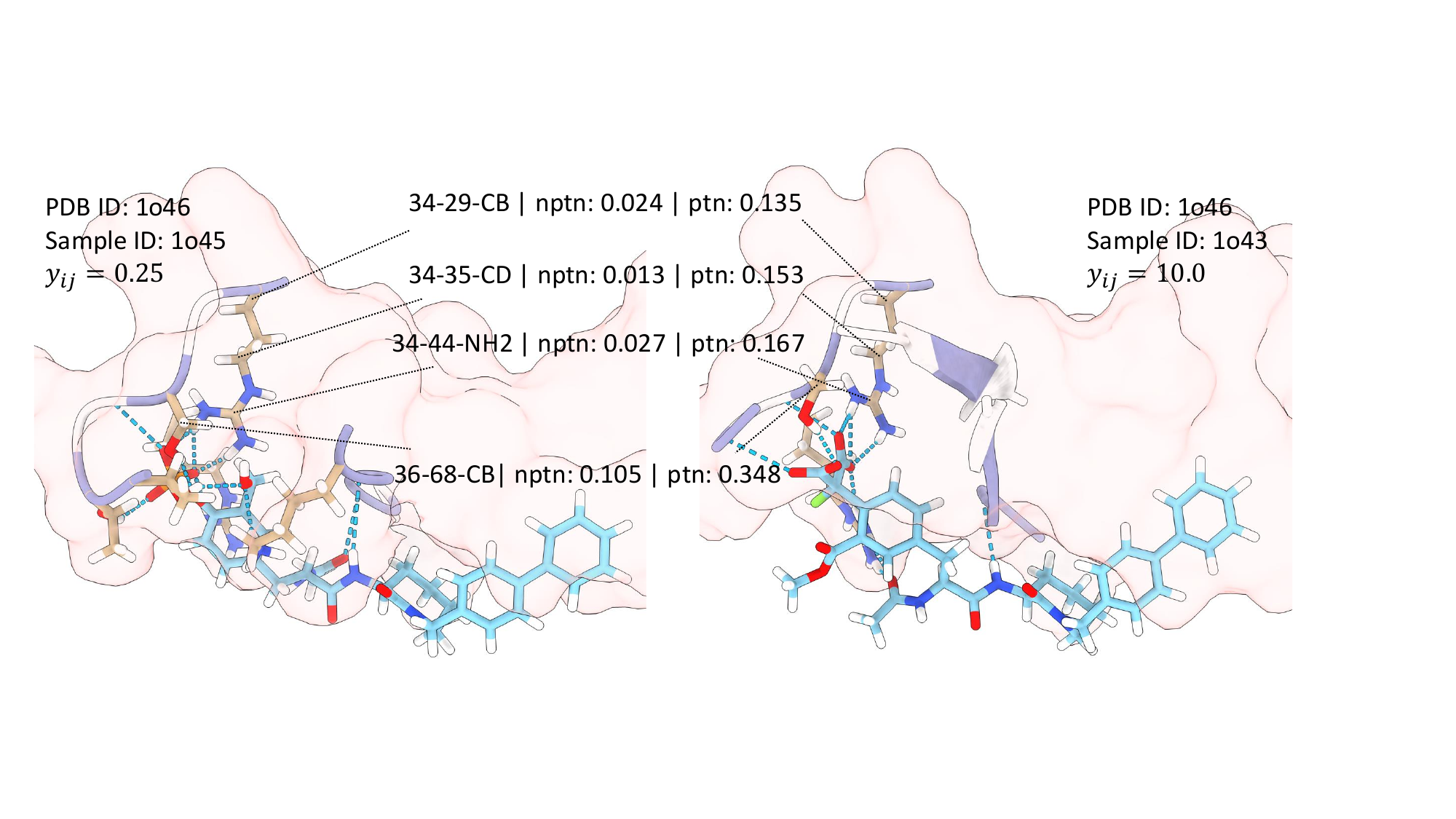}
    }\vspace{-0.5em}
    {
    \includegraphics[width=0.49\textwidth, trim=80 90 50 120, clip]{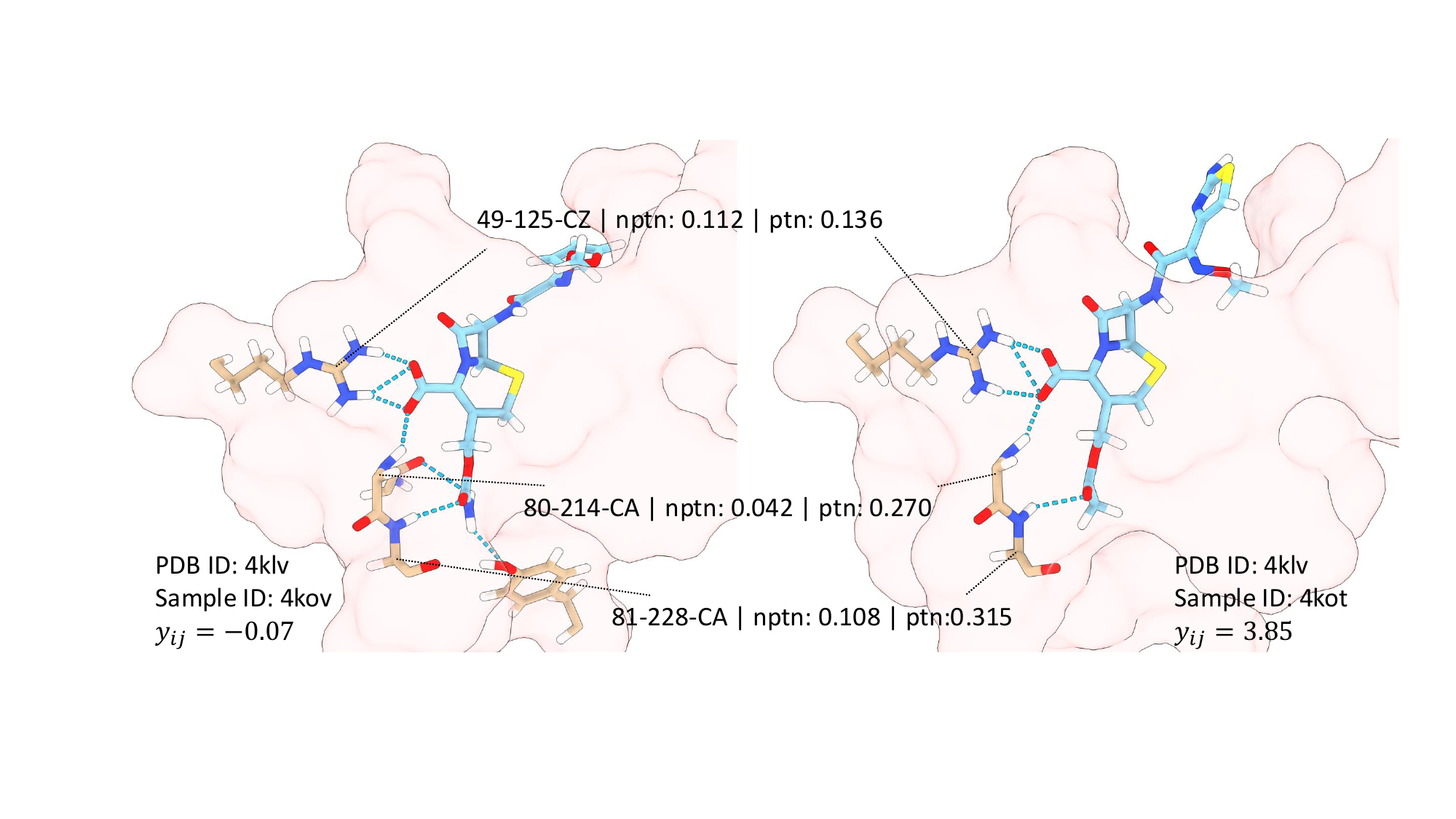}
    }
    \caption{Representation similarity of `key atoms' from non-pretrained and pretrained ECBind.}\vspace{-1em}\label{fig:keynodes}
    \label{fig:enter-label}\vspace{-1em}
\end{figure}
\vspace{-0.2em}
 To figure out what contributes to the performance gain, we suppose and analyze two factors in pretraining in detail. Besides, an ablation study is provided in Appendix.~\ref{app:ablation} on hyper-parameter sensitivity, and Appendix.~\ref{app:time_memory}  gives cost of inference time and training memory. 

\textbf{Accurate reconstruction of electron cloud helps.} We hypothesize that the relatively modest performance gains on the LBA dataset may stem from factors related to dataset size and quality. On one hand, the MISATO dataset is over twice the size of the LBA dataset. On the other hand, MISATO benefits from noise reduction through quantum chemistry calculations, molecular dynamics simulations, and refinements such as hydrogen atom adjustments and coordinate corrections, resulting in more stable and reliable protein-ligand structures as well as electron cloud information. In contrast, the LBA dataset suffers from noise in the original atomic coordinates, hydrogen atom deposits, and consequently unreliable electron cloud data. These issues make it challenging for the pretrained tokenizer to capture meaningful patterns in both the electron cloud and atomic structures.
We also observed similar trends during training (See Figure.~\ref{fig:lossec}\&\ref{fig:lossfa}): under identical tokenizer architectures, the model consistently achieved lower losses in both electron cloud reconstruction loss ($L^{\mathsf{ec}}$) and atomic topology/structure-related losses on MISATO, with test $L^{\mathsf{ec}}$ values being as low as a quarter of those observed on LBA. 
To further investigate, we separately establish two more variants of ECBind, which are fine-tuned only with Full-Atom Tokenizer (Sec.~\ref{sec:fatoken}) and the Electron-Cloud Tokenizer (Sec.~\ref{sec:ectoken}) on these two tasks. From Figure.~\ref{fig:codeablation}, for MISATO, the two codebooks exhibited a synergistic effect, with the Electron-Cloud codebooks contributing more performance gains compared to the Full-Atom codebooks. However, for LBA, using only the Full-Atom Tokenizer was sufficient to achieve performance comparable to ECBind-stdt. This indicates that imprecise electron-cloud information provides little additional benefit when the model can already effectively capture atomic-level structures in LBA.

\textbf{Effective perception of structural difference helps.} Secondly, we aim to understand why the pretrained ECBind shows such significant improvements in per-structure evaluations compared to its non-pretrained counterpart. An observation is that in a binding site, there are typically over 500 atoms from the protein and fewer than 50 from the molecule. Per-structure evaluations are conducted within the same protein binding to different molecules. We hypothesize that pretraining on both atom-level and electron-cloud-level structures enables the model to capture subtle structural differences from changes in molecules, leading to more accurate predictions. In contrast, the non-pretrained ECBind, while performing well on overall evaluations, primarily focuses on protein structures. Its strong overall performance likely stems from correctly distinguishing protein structural differences rather than capturing the intermolecular interaction patterns within the binding site.
To investigate this, we select two different molecules binding to the same protein and define the `key interacting nodes' as atom pairs from two biomolecules within a distance of $\leq 4\ \text{\AA}$. We extract the representations of proteins' key nodes before passing them through the decoder. Since we employed a spherical projection codebook, the representation similarity between the selected key nodes of the two binding sites is computed using cosine similarity. As shown in Figure.~\ref{fig:keynodes}, the atom representations after tokenization exhibit greater distinctiveness, preventing the encoder from over-smoothing them and the decoder thus from confusing them, resulting in more accurate predictions.

\vspace{-0.5em}
\section{Conclusion and Limitation}
We propose ECBind which integrates electronic and atomic structures for protein-ligand interaction learning. 
% By employing tokenization techniques during pretraining, we address the redundancy of electron cloud signals. Additionally, we develop an electron-cloud-agnostic student model to enhance applicability across diverse scenarios.
One limitation of ECBind is the additional computational cost associated with pretraining, which often relies on high-quality binding complex structures to achieve significant performance gains. Moreover, ECBind currently lacks fragment-level knowledge, and incorporating this information could further improve its predictive accuracy. Exploring these aspects will be our future work.

\section*{Broader Impacts}
By introducing quantum-scale electron cloud information into molecular representation learning, ECBind paves the way for a deeper and more physically grounded understanding of biomolecular interactions. Its ability to capture fine-grained electronic features provides a new method of precise prediction of protein-ligand binding affinity, potentially accelerating drug discovery processes. Beyond drug design, ECBind’s approach could be extended to diverse applications, such as modeling enzyme catalysis, understanding allosteric regulation, or designing functional molecular materials with targeted reactivity profiles.
Furthermore, the knowledge distillation component of ECBind ensures accessibility and efficiency, making it feasible to deploy advanced quantum-informed models in large-scale virtual screening and real-time therapeutic evaluation pipelines. This democratization of quantum-level insights—without prohibitive computational costs—may foster broader adoption of physically-aware AI in biology and chemistry, catalyzing new interdisciplinary research and innovation.

\bibliography{ref}
\bibliographystyle{plainnat}
%%%%%%%%%%%%%%%%%%%%%%%%%%%%%%%%%%%%%%%%%%%%%%%%%%%%%%%%%%%%

% \input{checklist}
\newpage
\appendix

\section{Gallery of Notations} 

\begin{table}[h]
\centering \resizebox{0.68\linewidth}{!}{
\begin{tabular}{|c|c|}
\hline
\textbf{Symbol} & \textbf{Description} \\
\hline
$\mathcal{B}$ &  Protein-ligand pair \\
$\mathcal{P}$ &  Protein atomic structures \\
$\mathcal{M}$ & Molecule~(ligand) atomic structures \\
$\bm{V}$ &  Node set of $\mathcal{P}$ and $\mathcal{M}$ \\
$\bm{E}$ &  Edge set of $\mathcal{P}$ and $\mathcal{M}$ \\
$N_{\mathrm{fa}}$ &  Number of atoms \\
$N_\mathrm{ec}$ & Number of electron cloud signals\\
$\mathcal{I}_{\mathrm{rec}}$ & Receptors' index set \\
$\mathcal{I}_{\mathrm{lig}}$ & Ligands' index set \\
$a_i$ & Atom types \\
$\bm{\mathrm{x}}^{\mathsf{a}}_i$ & 3D position of atom $a_i$ \\
$\mathcal{D}$ & $\{(u_i, \bm{\mathrm{x}}^\mathsf{u}_i)\}_{1 \leq i \leq N_{\mathrm{ec}}}$, Electron cloud signals \\
$\bm{\mathrm{x}}^{\mathsf{u}}_i$ & Eletron density's position \\
$u_i$ & Electron density at position $\bm{\mathrm{x}}^{\mathsf{u}}_i$ \\
$\bm{z}^{\mathsf{u}}_i$ & High-dimensional embedding extracted by $u_i$ \\
$\bm{z}^{\mathsf{a}}_i$ & High-dimensional embedding extracted by $a_i$ \\
$\bm{Z}^{\mathsf{u}}_i$ & $(\bm{z}^{\mathsf{a}}_i, \bm{z}^{\mathsf{u}}_{j_1}, \ldots, \bm{z}^{\mathsf{u}}_{j_K})$, embedding sequence of patches. \\
$\bm{X}^{\mathsf{u}}_i$ & $(\bm{\mathrm{x}}^{\mathsf{a}}_i, \bm{\mathrm{x}}^{\mathsf{u}}_{j_1}, \ldots, \bm{\mathrm{x}}^{\mathsf{u}}_{j_K})$, the corresponding position sequence. \\
$\gamma_\mathrm{pt}$ & Predefined smoothing coefficient \\
$\bm{Q}^{\mathsf{iv}}$ & Invariant-feature queries \\
$\bm{Q}^{\mathsf{ge}}$ & Geometric-feature queries \\
$\bm{K}^{\mathsf{iv}}$ & Invariant-queries keys \\
$\bm{K}^{\mathsf{ge}}$ & The geometric-queries keys \\
$\bm{V}$ &Values in attention\\
$H$ & Number of attention heads \\
$\bm{\mathrm{A}}^{\mathsf{iv}}$ & Classical attention weight matrix \\
$\bm{\mathrm{A}}^{\mathsf{ge}}$ & Geometric attention weight matrix \\
$\bm{\mathrm{A}}$ & Final attention weight matrix \\
$\mathrm{pdist}$ & Pairwise-distance function \\
$\bm{O}$ & GeoMHA-based transformer's output \\
$\phi^{\mathsf{ec}}(\cdot, \cdot)$ & GeoMHA-based transformer on electron-cloud-level message passing\\
$\phi^{\mathsf{fa}}(\cdot, \cdot)$ & Full-atom GeoMHA-based transformer \\
$\mathcal{A}^\mathsf{u(1)}, \mathcal{A}^\mathsf{u(2)}$ & Codebook for 3D electronic structures and 2D atomic attributes \\
$\bm{h}^\mathsf{u}_i$ & Patchwise representation \\
$\bm{h}_i'^{\mathsf{u}}$ & Discrete codes of $\bm{h}^\mathsf{u}_i$ \\
$L^{\mathsf{ec}}$ & The loss function of electron \\
$L^{\mathsf{at}}$ & The loss function of atom \\
$L^{\mathsf{st}}$ & The intra-molecular binned distance classification loss \\
$L^{\mathsf{it}}$ & The inter-molecular interaction loss \\
$L^{\mathsf{cmt}}$ & The commitment loss \\
$L^{\mathsf{u}}$ & The training objective for the tokenizer \\
$\alpha^{\mathsf{u}}$ & The loss weight \\
\hline
\end{tabular}}
\caption{Gallery of Notations}
\end{table}
\section{Proof of Invariance} \label{app:invproof}
\textbf{Lemma}.
For the feature $\bm{H} \in \mathbb{R}^{N \times D}$ which is invariant to the translation, rotation and reflection,
then for any function $F: \mathbb{R}^{N \times D} \rightarrow \mathbb{R}^{N \times D'}$, such that  $\bm{H}' = F(\bm{H}) \in \mathbb{R}^{N \times D'}$ will be invariant to the corresponding operations.

Then the invariance of $\bm{O} = \bm{AV}$ should prove that $\bm{A}$ and $\bm{V}$ are invariance to geometric translation.
In this way, since $\bm{V}$ is the representation after linear transformation of $\bm{Z}$, which is the atom type embedding, it is obvious that $\bm{V}$ is invariant.
For the attention matrix $\bm{A} = \text{softmax}(\bm{A}^{\mathsf{iv}} - \bm{A}^{\mathsf{ge}})$, $\bm{A}^{\mathsf{iv}}$ is the direct product of two invariant feature, and $\bm{A}^{\mathsf{ge}}$ is to measure the distance, which in detail, for the rotation matrix $\bm{\mathrm{R}}$: 
\begin{equation}
\begin{aligned}
    &\frac{\text{softplus}(w)}{\sqrt{3}} \mathrm{pdist} (\bm{\mathrm{R}}\bm{X} \cdot \bm{Q},\bm{\mathrm{R}}\bm{X} \cdot \bm{K})  \\
    =& \frac{\text{softplus}(w)}{\sqrt{3}} \mathrm{pdist} (\bm{X} \cdot \bm{Q},\bm{X} \cdot \bm{K}) \\
    =& \bm{A}^{\mathsf{ge}}
\end{aligned}
\end{equation}
because the pairwise distance function is invariant to rotation. 
As we apply the zero-center-of-mass technique, and $\bm{\mathrm{R}}$ is the rotation matrix satisfies $\bm{\mathrm{R}} \bm{\mathrm{R}}^{\mathsf{T}} = \bm{\mathrm{I}}$, the translation vector $\bm{\mathrm{t}} = \bm{0}$.
Therefore, $\bm{A}^{\mathsf{ge}}$ is invariant to the rotation and translation, leading to the invariance of $\bm{A}$.

\section{Tokenizing Atom-level Structure}
\subsection{Full-atom Tokenizer} 
\label{app:structtoken}
The macro-environment around the $i$-th atom is defined similarly to the patchification of electron clouds. The full-atom GeoMHA-based transformer $\phi^{\mathsf{fa}}(\cdot, \cdot)$ encode the macro-environment as
\begin{equation}
    \begin{aligned}
        \quad\quad \quad \bm{h}^\mathsf{a}_i &= [\phi^{\mathsf{fa}}(\bm{Z}^{\mathsf{a}}_i, \bm{X}^{\mathsf{a}}_i)]_1, \quad&\text{where } \\
        \bm{Z}^{\mathsf{a}}_i &= (\bm{z}_i^{\mathsf{a}}, \bm{z}_{i_1}^{\mathsf{a}}, \ldots, \bm{z}_{i_K}^{\mathsf{a}}), \quad\quad\quad\quad &\text{and } \\
        \bm{X}^{\mathsf{a}}_i &= (\bm{\mathrm{x}}_i^{\mathsf{a}}, \bm{\mathrm{x}}_{i_1}^{\mathsf{a}}, \ldots, \bm{\mathrm{x}}_{i_K}^{\mathsf{a}}).
    \end{aligned}
\end{equation}
Here, the $i_k$-th atom belongs to the neighborhood of the $i$-th atom based on their Euclidean distance.

Two codebooks, $\mathcal{A}^{\mathsf{a}} = \mathcal{A}^{\mathsf{a}(1)} \times \mathcal{A}^{\mathsf{a}(2)}$, are constructed to capture the 3D conformational geometry and 2D atomic topology, following the same approach as Eq.~\ref{eq:vqu}. For $\bm{h}_i'^{\mathsf{a}(2)}$ obtained from $\mathcal{A}^{\mathsf{a}(2)}$, the pretraining follows the Masked AutoEncoder (MAE) tasks described in Sec.~\ref{sec:ecrec}, leading to the loss function $L^{\mathsf{at}}$. For $\bm{h}_i'^{\mathsf{a}(1)}$, we randomly mask the positions of a subset of atoms in the input and reconstruct the geometries, as discussed in the following sections.

\subsection{Structure Decoder and Loss Function.} 
\label{app:structdec}
A classical transformer is employed to decode the tokenized representation $\bm{h}_i'^{\mathsf{a}(1)}$ into a structural one, defined as $\bm{h}_i^{\mathsf{st}} = \psi^{\mathsf{st}}(\bm{h}_i'^{\mathsf{a}(1)})$.

Two types of loss functions are used: (i) intra-molecular binned distance classification loss and (ii) inter-molecular interaction loss.

For (i), we first compute the ground-truth Euclidean distance between atoms as $\|\bm{\mathrm{x}}_i^{\mathsf{a}} - \bm{\mathrm{x}}_j^{\mathsf{a}}\|_2$. This distance is divided into discrete intervals $[0, 2, 3, 4, \dots, 20, +\infty]$, resulting in a pairwise distance label $y^{\mathsf{st}}_{i,j}$. 
To generate pairwise logits for structure prediction, we employ a \texttt{pairwise\_proj\_head}, which processes input embeddings $\bm{H} \in \mathbb{R}^{N \times D}$ and outputs logits $\bm{O} = \text{logit} \in \mathbb{R}^{L \times L \times d'}$. The method follows these steps:

\begin{algorithm}
    \caption{\texttt{pairwise\_proj\_head}}
    \label{alg:pairwise_proj_head}
    \begin{algorithmic}[1]
        \REQUIRE $\bm{H}  \in \mathbb{R}^{L \times d}$
        \STATE $\bm{Q}, \bm{K} \gets \text{Linear}(\bm{H} ), \text{Linear}(\bm{H} )$
        \STATE $\text{prod}_{i,j,:}, \text{diff}_{i,j,:} \gets \bm{Q}_{j,:} \odot \bm{K}_{i,:} , \bm{Q}_{j,:} -\bm{K}_{i,:}$
        \STATE $\bm{O} \gets \text{MLP}([\text{prod} ; \text{diff}]) \in \mathbb{R}^{L \times L \times d'}$
        \OUTPUT $\bm{O}$
    \end{algorithmic}
\end{algorithm}

Therefore, by employing the discussed pair-wise head,
we map the pair representations $(\bm{h}_i^{\mathsf{st}}, \bm{h}_j^{\mathsf{st}})$ to logits $\mathrm{logit}_{i,j}^{\mathsf{st}} \in \mathbb{R}^{20}$. Cross-entropy is applied for $ i,j \in \mathcal{I}_\mathrm{rec} \text{ or }  i,j \in \mathcal{I}_\mathrm{lig}$ as the loss function:
\begin{equation}
    \begin{aligned}
        L^{\mathsf{st}} = \mathrm{CE}_{i,j}(\mathrm{logit}_{i,j}^{\mathsf{st}}, y^{\mathsf{st}}_{i,j}).
    \end{aligned}
\end{equation}

For (ii), a binary interaction label $y^{\mathsf{it}}_{i,j}$ is defined, indicating whether two atoms are interacting based on their distance ( $\leq$ 6\AA). Another pairwise head maps $(\bm{h}_i^{\mathsf{st}}, \bm{h}_j^{\mathsf{st}})$ to $\text{prob}_{i,j}^{\mathsf{it}}$, representing the probability of interaction. Binary cross-entropy is used as the loss function:
\begin{equation}
    \begin{aligned}
        L^{\mathsf{it}} = \mathrm{BCE}_{i,j} (\text{prob}_{i,j}^{\mathsf{it}}, y^{\mathsf{it}}_{i,j}),
    \end{aligned}
\end{equation}
where $i \in \mathcal{I}_\mathrm{rec}, j \in \mathcal{I}_\mathrm{lig}$ or $i \in \mathcal{I}_\mathrm{lig}, j \in \mathcal{I}_\mathrm{rec}$.

%%%%%%%%%%%%%%%%%%%%%%%%%%%%%%%%%%%%%%%%%%%%%%%%%%%%%%%%%%%%

\section{Baseline Method Introduction} \label{app:methodintro}
In this section, we provide a brief overview of the baseline methods selected for comparison in this study, as listed:
\begin{itemize}
    \item SchNet is a deep learning architecture designed for modeling quantum interactions in molecules using continuous-filter convolutional layers. It predicts total energy and interatomic forces while adhering to quantum-chemical principles, such as rotational invariance and smooth potential energy surfaces.
    \item EGNN is a 3D equivariant graph neural network that not only maintains performance but also avoids the complex computations of higher-order representations in intermediate layers, extending equivariant properties to dimensions beyond 3D.
    \item DimeNet is a graph neural network that enhances quantum mechanical property predictions by incorporating directional message passing. Unlike traditional models that use only atomic distances, DimeNet embeds directional information between atoms, ensuring rotational equivariance. It uses spherical Bessel functions and spherical harmonics for efficient, orthogonal representations, outperforming Gaussian radial basis functions while using fewer parameters.
    \item GemNet is a geometric message-passing neural network that uses directed edge embeddings and two-hop message passing, making it a universal approximator for predictions that are invariant to translation and equivariant to permutation and rotation.
    \item TorchMD is a framework designed for molecular simulations that combines classical and machine learning-based potentials. It represents all force computations, such as bond, angle, and Coulomb interactions, using PyTorch arrays and operations, enabling seamless integration with neural network potentials. The framework supports learning and simulating both ab initio and coarse-grained models, and has been validated through standard Amber all-atom simulations and protein folding tasks. 
    \item MACE is an equivariant message passing neural network (MPNN) model that utilizes higher-body order messages, specifically four-body messages, which reduces the number of required message passing iterations to just two, making it fast and highly parallelizable.
    \item Equiformer is a graph neural network that extends Transformer architectures to 3D atomistic graphs by incorporating SE(3)/E(3)-equivariant features using irreducible representations (irreps). It improves the original Transformer model by replacing standard operations with their equivariant counterparts and integrating tensor products, enabling efficient encoding of equivariant information. Additionally, Equiformer introduces a novel equivariant graph attention mechanism, which enhances performance by using multi-layer perceptron attention and non-linear message passing.
    \item LEFTNet is a geometric GNN that incorporates local substructure encoding (LSE) and frame transition encoding (FTE) to effectively represent global geometric information from local patches, achieving state-of-the-art performance on molecular property prediction tasks.
    \item HEGNN is a high-degree version of EGNN to increase the expressivity by incorporating high-degree steerable vectors while still maintaining EGNN’s advantage through the scalarization trick.
\end{itemize}       
\section{Training Detail} \label{app:training}
Experiments are conducted based on Pytorch 2.4.2 on a hardware platform with Intel(R) Xeon(R) Gold 6240R @ 2.40GHz CPU and NVIDIA A100 GPU. 
We give the detailed parameters for training the included models in Table.~\ref{tab:baselineparam1}, ~\ref{tab:baselineparam2} and ~\ref{tab:hyperparam3}.
Besides, the training epoch for LBA and LEP, MISATO is 20, while the finetuning for LBA is 10 and for MISATO and LEP is 20.
\section{Ablation Study} \label{app:ablation}

We here aim to figure out the effects of hyper-parameters, including (i) codebook size, which is $|\mathcal{A^{\mathsf{u(1)}}}| = |\mathcal{A^{\mathsf{u(2)}}}|$ , (ii) embedding size, which is $D$, (iii) layer number of encoder $N_{\mathrm{enc}}$, (iv) layer number of decoder $N_{\mathrm{dec}}$,
% \begin{table}[] \centering
% \caption{The ablation study. PS means per-structure. }
% \begin{tabular}{
% >{\columncolor[HTML]{D4D4D4}}rrr}
% \toprule
% \textbf{D}   & Pearson\_PS & Spearman\_PS \\
% \midrule
% \textbf{32}  & 0.226       & 0.209        \\
% \textbf{64}  & 0.314       & 0.297        \\
% \textbf{128} & 0.423       & 0.448        \\
% \textbf{256} & 0.431       & 0.445       \\
% \bottomrule
% \end{tabular} \hspace{ 3em }
% \begin{tabular}{
% >{\columncolor[HTML]{D4D4D4}}rrr}
% \toprule
% \textbf{codebook\_size} & Pearson\_PS & Spearman\_PS \\
% \midrule
% \textbf{256}            & 0.378       & 0.366        \\
% \textbf{512}            & 0.389       & 0.392        \\
% \textbf{1024}           & 0.454       & 0.433        \\
% \textbf{2048}           & 0.431       & 0.445       \\
% \bottomrule
% \end{tabular}
% \begin{tabular}{
% >{\columncolor[HTML]{D4D4D4}}rrr}
% \toprule
% \textbf{n\_layer\_enc} & Pearson\_PS & Spearman\_PS \\
% \midrule
% \textbf{32}            & 0.226       & 0.209        \\
% \textbf{64}            & 0.314       & 0.297        \\
% \textbf{128}           & 0.423       & 0.448        \\
% \textbf{256}           & 0.431       & 0.445      \\
% \bottomrule
% \end{tabular}\hspace{3em}
% \begin{tabular}{
% >{\columncolor[HTML]{D4D4D4}}rrr}
% \toprule
% \textbf{n\_layer\_dec} & Pearson\_PS & Spearman\_PS \\
% \midrule
% \textbf{1}             & 0.245       & 0.248        \\
% \textbf{2}             & 0.287       & 0.305        \\
% \textbf{3}             & 0.431       & 0.445        \\
% \textbf{6}             & 0.392       & 0.410       \\
% \bottomrule
% \end{tabular}

% \end{table}
\begin{table}[h]
\centering
\caption{The ablation study. PS means per-structure.}
\label{tab:ablation}
\renewcommand{\arraystretch}{1.1}

% 第一行：D 和 codebook_size
\begin{minipage}[t]{0.45\textwidth}
\centering
\begin{tabular}{
>{\centering\arraybackslash}p{1.8cm}
>{\centering\arraybackslash}p{1.8cm}
>{\centering\arraybackslash}p{1.8cm}}
\toprule
\textbf{D} & Pearson\_PS & Spearman\_PS \\
\midrule
\textbf{32}  & 0.226 & 0.209 \\
\textbf{64}  & 0.314 & 0.297 \\
\textbf{128} & 0.423 & 0.448 \\
\textbf{256} & 0.431 & 0.445 \\
\bottomrule
\end{tabular}
\end{minipage}
\hspace{1em}
\begin{minipage}[t]{0.45\textwidth}
\centering
\begin{tabular}{
>{\centering\arraybackslash}p{1.8cm}
>{\centering\arraybackslash}p{1.8cm}
>{\centering\arraybackslash}p{1.8cm}}
\toprule
\textbf{codebook\_size} & Pearson\_PS & Spearman\_PS \\
\midrule
\textbf{256}  & 0.378 & 0.366 \\
\textbf{512}  & 0.389 & 0.392 \\
\textbf{1024} & 0.454 & 0.433 \\
\textbf{2048} & 0.431 & 0.445 \\
\bottomrule
\end{tabular}
\end{minipage}
\\[1em]
% 第二行：n_layer_enc 和 n_layer_dec
\begin{minipage}[t]{0.45\textwidth}
\centering
\begin{tabular}{
>{\centering\arraybackslash}p{1.8cm}
>{\centering\arraybackslash}p{1.8cm}
>{\centering\arraybackslash}p{1.8cm}}
\toprule
\textbf{n\_layer\_enc} & Pearson\_PS & Spearman\_PS \\
\midrule
\textbf{32}  & 0.226 & 0.209 \\
\textbf{64}  & 0.314 & 0.297 \\
\textbf{128} & 0.423 & 0.448 \\
\textbf{256} & 0.431 & 0.445 \\
\bottomrule
\end{tabular}
\end{minipage}
\hspace{1em}
\begin{minipage}[t]{0.45\textwidth}
\centering
\begin{tabular}{
>{\centering\arraybackslash}p{1.8cm}
>{\centering\arraybackslash}p{1.8cm}
>{\centering\arraybackslash}p{1.8cm}}
\toprule
\textbf{n\_layer\_dec} & Pearson\_PS & Spearman\_PS \\
\midrule
\textbf{1} & 0.245 & 0.248 \\
\textbf{2} & 0.287 & 0.305 \\
\textbf{3} & 0.431 & 0.445 \\
\textbf{6} & 0.392 & 0.410 \\
\bottomrule
\end{tabular}
\end{minipage}

\end{table}

\begin{table}[h]
\centering
\caption{Hyperparameters for each baseline on each task.(Part 1)}
\label{tab:baselineparam1}
\scalebox{0.8}{
\begin{tabular}{llllllll}
\toprule
hyperparameter       & MISATO      & LBA       & LEP       & hyperparameter       & MISATO      & LBA       & LEP       \\ \hline
\multicolumn{8}{c}{SchNet-fragment}                                                                      \\ \hline
$d_h$                & 64      & 64        & 128        & max\_n\_vertex       & 2000     & 2000      & 500      \\
lr                   & $10^{-3}$& $10^{-3}$ & $10^{-3}$ & final\_lr     & $10^{-6}$& $10^{-6}$ & $10^{-4}$ \\
max\_epoch           & 10       & 10        & 70        & save\_topk           & 3        & 3         & 5         \\ \hline
\multicolumn{8}{c}{SchNet-atom}                                                                      \\ \hline
$d_h$                & 64      & 64        & 128        & max\_n\_vertex       & 2000     & 2000      & 1500      \\
lr                   & $10^{-3}$& $10^{-3}$ & $5\times10^{-4}$ & final\_lr     & $10^{-6}$& $10^{-6}$ & $10^{-4}$ \\
max\_epoch           & 10       & 10        & 90        & save\_topk           & 3        & 3         & 7         \\ \hline
\multicolumn{8}{c}{SchNet-hierarchical}                                                                      \\ \hline
$d_h$                & 64      & 64        & 128        & max\_n\_vertex       & 2000     & 2000      & 500      \\
lr                   & $10^{-3}$& $10^{-3}$ & $10^{-3}$ & final\_lr     & $10^{-6}$& $10^{-6}$ & $10^{-4}$ \\
max\_epoch           & 10       & 10        & 70        & save\_topk           & 3        & 3         & 5         \\ \hline
\multicolumn{8}{c}{DimeNet++-fragement}                                                                   \\ \hline
$d_h$                & 64      & 64        & 128        & max\_n\_vertex       & 2000     & 2000      & 500      \\
lr                   & $10^{-3}$& $10^{-3}$ & $10^{-3}$ & final\_lr     & $10^{-6}$& $10^{-6}$ & $10^{-4}$ \\
max\_epoch           & 10       & 10        & 70        & save\_topk           & 3        & 3         & 5         \\ \hline
\multicolumn{8}{c}{DimeNet++-atom}                                                                   \\ \hline
$d_h$                & 64      & 64        & 128        & max\_n\_vertex       & 2000     & 2000      & 1500      \\
lr                   & $10^{-3}$& $10^{-3}$ & $5\times10^{-4}$ & final\_lr     & $10^{-6}$& $10^{-6}$ & $10^{-4}$ \\
max\_epoch           & 10       & 10        & 90        & save\_topk           & 3        & 3         & 7         \\ \hline
\multicolumn{8}{c}{DimeNet++-hierarchical}                                                                   \\ \hline
$d_h$                & 64      & 64        & 128        & max\_n\_vertex       & 2000     & 2000      & 500      \\
lr                   & $10^{-3}$& $10^{-3}$ & $10^{-3}$ & final\_lr     & $10^{-6}$& $10^{-6}$ & $10^{-4}$ \\
max\_epoch           & 10       & 10        & 70        & save\_topk           & 3        & 3         & 5         \\ \hline
\multicolumn{8}{c}{EGNN-fragment}                                                                        \\ \hline
$d_h$                & 64      & 64        & 128        & max\_n\_vertex       & 2000     & 2000      & 500      \\
lr                   & $10^{-3}$& $10^{-3}$ & $10^{-3}$ & final\_lr     & $10^{-6}$& $10^{-6}$ & $10^{-4}$ \\
max\_epoch           & 10       & 10        & 70        & save\_topk           & 3        & 3         & 5         \\ \hline
\multicolumn{8}{c}{EGNN-atom}                                                                        \\ \hline
$d_h$                & 64      & 64        & 128        & max\_n\_vertex       & 2000     & 2000      & 1500      \\
lr                   & $10^{-3}$& $10^{-3}$ & $5\times10^{-4}$ & final\_lr     & $10^{-6}$& $10^{-6}$ & $10^{-4}$ \\
max\_epoch           & 10       & 10        & 90        & save\_topk           & 3        & 3         & 7         \\ \hline
\multicolumn{8}{c}{EGNN-hierarchical}                                                                        \\ \hline
$d_h$                & 64      & 64        & 128        & max\_n\_vertex       & 2000     & 2000      & 500      \\
lr                   & $10^{-3}$& $10^{-3}$ & $10^{-3}$ & final\_lr     & $10^{-6}$& $10^{-6}$ & $10^{-4}$ \\
max\_epoch           & 10       & 10        & 70        & save\_topk           & 3        & 3         & 5         \\ \hline
\multicolumn{8}{c}{ET-fragment}                                                                          \\ \hline
$d_h$                & 64      & 64        & 128        & max\_n\_vertex       & 2000     & 2000       & 500      \\
lr                   & $10^{-3}$& $10^{-3}$ & $10^{-3}$ & final\_lr     & $10^{-6}$& $10^{-6}$ & $10^{-4}$ \\
max\_epoch           & 10       & 10        & 70        & save\_topk           & 3        & 3         & 5         \\ \hline
\multicolumn{8}{c}{ET-atom}                                                                          \\ \hline
$d_h$                & 64      & 64        & 128        & max\_n\_vertex       & 2000     & 2000       & 1500      \\
lr                   & $10^{-3}$& $10^{-3}$ & $5\times10^{-4}$ & final\_lr     & $10^{-6}$& $10^{-6}$ & $10^{-4}$ \\
max\_epoch           & 10       & 10        & 90        & save\_topk           & 3        & 3         & 7         \\ \hline
\multicolumn{8}{c}{ET-hierarchical}                                                                          \\ \hline
$d_h$                & 64      & 64        & 128        & max\_n\_vertex       & 2000     & 2000       & 500      \\
lr                   & $10^{-3}$& $10^{-3}$ & $10^{-3}$ & final\_lr     & $10^{-6}$& $10^{-6}$ & $10^{-4}$ \\
max\_epoch           & 10       & 10        & 70        & save\_topk           & 3        & 3         & 5         \\ \hline
\multicolumn{8}{c}{GemNet-fragment}                                                                      \\ \hline
$d_h$                & 64      & 64        & 128         & max\_n\_vertex       & 1500     & 2000      & 500         \\
lr                   & $10^{-3}$& $10^{-3}$& $10^{-3}$  & final\_lr      & $10^{-6}$& $10^{-6}$ & $10^{-4}$     \\
max\_epoch           & 10       & 10        & 70         & save\_topk           & 3        & 3         & 5         \\ \hline
\multicolumn{8}{c}{GemNet-atom}                                                                      \\ \hline
$d_h$                & 32      & 64        & 128         & max\_n\_vertex       & 200     & 500      & 1500         \\
lr                   & $10^{-3}$& $10^{-3}$& $5\times10^{-4}$  & final\_lr      & $10^{-6}$& $10^{-6}$ & $10^{-4}$     \\
max\_epoch           & 10       & 10        & 90         & save\_topk           & 3        & 3         & 7         \\ \hline
\multicolumn{8}{c}{GemNet-hierarchical}                                                                      \\ \hline
$d_h$                & 32      & 64        & 128         & max\_n\_vertex       & 200     & 500      & 500         \\
lr                   & $10^{-3}$& $10^{-3}$& $10^{-3}$  & final\_lr      & $10^{-6}$& $10^{-6}$ & $10^{-4}$     \\
max\_epoch           & 10       & 10        & 70         & save\_topk           & 3        & 3         & 5         \\ \hline
\end{tabular}
}
\vskip -0.2in
\end{table}

\begin{table}[h]
\centering
\caption{Hyperparameters for each baseline on each task.(Part 2)}
\label{tab:baselineparam2}
\scalebox{0.8}{
\begin{tabular}{llllllll}
\toprule
hyperparameter       & MISATO      & LBA       & LEP       & hyperparameter       & MISATO      & LBA       & LEP       \\ \hline
\multicolumn{8}{c}{MACE-fragment}                                                                      \\ \hline
$d_h$                & 64      & 64        & 128         & max\_n\_vertex       & 1000     & 2000      & 500         \\
lr                   & $10^{-3}$& $10^{-3}$& $10^{-3}$  & final\_lr      & $10^{-6}$& $10^{-6}$ & $10^{-4}$     \\
max\_epoch           & 10       & 10        & 70         & save\_topk           & 3        & 3         & 5         \\ \hline
\multicolumn{8}{c}{MACE-atom}                                                                      \\ \hline
$d_h$                & 64      & 64        & 128         & max\_n\_vertex       & 2000     & 2000      & 1500         \\
lr                   & $10^{-3}$& $10^{-3}$& $5\times10^{-4}$  & final\_lr      & $10^{-6}$& $10^{-6}$ & $10^{-4}$     \\
max\_epoch           & 10       & 10        & 90         & save\_topk           & 3        & 3         & 7         \\ \hline
\multicolumn{8}{c}{MACE-hierarchical}                                                                      \\ \hline
$d_h$                & 64      & 64        & 128         & max\_n\_vertex       & 2000     & 2000      & 500         \\
lr                   & $10^{-3}$& $10^{-3}$& $10^{-3}$  & final\_lr      & $10^{-6}$& $10^{-6}$ & $10^{-4}$     \\
max\_epoch           & 10       & 10        & 70         & save\_topk           & 3        & 3         & 5         \\ \hline
\multicolumn{8}{c}{Equiformer-fragment}                                                                      \\ \hline
$d_h$                & 32      & 64        & 128         & max\_n\_vertex       & 700     & 2000      & 500         \\
lr                   & $10^{-3}$& $10^{-3}$& $10^{-3}$  & final\_lr      & $10^{-6}$& $10^{-6}$ & $10^{-4}$     \\
max\_epoch           & 10       & 10        & 70         & save\_topk           & 3        & 3         & 5         \\ \hline
\multicolumn{8}{c}{Equiformer-atom}                                                                      \\ \hline
$d_h$                & -      & -       & -         & max\_n\_vertex       & -     & -      & -         \\
lr                   & -& -& -  & final\_lr      & -& - & -     \\
max\_epoch           & -       & -        & -         & save\_topk           & -        & -         & -         \\ \hline
\multicolumn{8}{c}{Equiformer-hierarchical}                                                                      \\ \hline
$d_h$                & -      & -        & -         & max\_n\_vertex       & -     & -      & -         \\
lr                   & -& -& -  & final\_lr      & -& - & -     \\
max\_epoch           & -       & -        & -         & save\_topk           & -        & -         & -         \\ \hline
\multicolumn{8}{c}{LEFTNet-fragment}                                                                      \\ \hline
$d_h$                & 64      & 64        & 128         & max\_n\_vertex       & 2000     & 2000      & 500         \\
lr                   & $10^{-3}$& $10^{-3}$& $10^{-3}$  & final\_lr      & $10^{-6}$& $10^{-6}$ & $10^{-4}$     \\
max\_epoch           & 10       & 10        & 70         & save\_topk           & 3        & 3         & 5         \\ \hline
\multicolumn{8}{c}{LEFTNet-atom}                                                                      \\ \hline
$d_h$                & 64      & 64        & 128         & max\_n\_vertex       & 2000     & 2000      & 1500         \\
lr                   & $10^{-3}$& $10^{-3}$& $5\times10^{-4}$  & final\_lr      & $10^{-6}$& $10^{-6}$ & $10^{-4}$     \\
max\_epoch           & 10       & 10        & 90         & save\_topk           & 3        & 3         & 7         \\ \hline
\multicolumn{8}{c}{LEFTNet-hierarchical}                                                                      \\ \hline
$d_h$                & 64      & 64        & 128         & max\_n\_vertex       & 2000     & 2000      & 500         \\
lr                   & $10^{-3}$& $10^{-3}$& $10^{-3}$  & final\_lr      & $10^{-6}$& $10^{-6}$ & $10^{-4}$     \\
max\_epoch           & 10       & 10        & 70         & save\_topk           & 3        & 3         & 5         \\ \hline
\multicolumn{8}{c}{GET-fragment}                                                                      \\ \hline
$d_h$                & 64      & 64        & 128         & max\_n\_vertex       & 2000     & 2000      & 500         \\
lr                   & $10^{-3}$& $10^{-3}$& $10^{-3}$  & final\_lr      & $10^{-6}$& $10^{-6}$ & $10^{-4}$     \\
max\_epoch           & 10       & 10        & 70         & save\_topk           & 3        & 3         & 5         \\ \hline
\multicolumn{8}{c}{GET-atom}                                                                      \\ \hline
$d_h$                & 64      & 64        & 128         & max\_n\_vertex       & 2000     & 2000      & 1500         \\
lr                   & $10^{-3}$& $10^{-3}$& $5\times10^{-4}$  & final\_lr      & $10^{-6}$& $10^{-6}$ & $10^{-4}$     \\
max\_epoch           & 10       & 10        & 90         & save\_topk           & 3        & 3         & 7         \\ \hline
\multicolumn{8}{c}{GET-hierarchical}                                                                      \\ \hline
$d_h$                & 64      & 64        & 128         & max\_n\_vertex       & 2000     & 2000      & 500         \\
lr                   & $10^{-3}$& $10^{-3}$& $10^{-3}$  & final\_lr      & $10^{-6}$& $10^{-6}$ & $10^{-4}$     \\
max\_epoch           & 10       & 10        & 70         & save\_topk           & 3        & 3         & 5         \\ \hline
\end{tabular}
}
\vskip -0.2in
\end{table}

\begin{table}[htbp]
\centering
\caption{Hyperparameters for our ECBind on each task.}
\label{tab:hyperparam3}
\scalebox{0.9}{
\begin{tabular}{llllllll}
\toprule
hyperparameter       & MISATO      & LBA       & LEP       & hyperparameter       & MISATO      & LBA       & LEP       \\ \hline
\multicolumn{8}{c}{ECBind-nptn}                                                                         \\ \hline
$D$                & 256      & 256        & 256        & $d_r$                & 16       & 32        & 64        \\
lr                   & $3\times10^{-5}$& $\times10^{-4}$ & $3\times 10^{-4}$ & final\_lr            & $10^{-6}$& $10^{-6}$ & $10^{-6}$ \\
max\_epoch           & 20       & 20        & 20        & save\_topk           & 3        & 3         & 3         \\
n\_layers\_enc            & 3        & 3         & 3         & n\_layers\_dec       & 3     & 3      & 3      \\
codebook\_size &2048&2048&2048& & & &\\
\hline
\end{tabular}
}
\vskip -0.2in
\end{table}

\section{Time and Memory Analysis} \label{app:time_memory}
We report the average inference time on the total test set and peak memory usage in the training of all models. The results are averaged over 3 independent runs using a single A100 GPU. Since our models are totally transformer-based which can integrate linear-attention mechanisms like FlashAttention, it is extremely fast in inference and much faster than graph-based models like EGNN and GemNet.

\begin{table}[h]
\centering
\caption{Comparison of inference time and memory consumption across models on MISATO.}
\begin{tabular}{lcc}
\toprule
\textbf{Model} & \textbf{Time (s)} & \textbf{Memory (MB)} \\
\midrule
EGNN-fragment           & 14.72   & 361.62 \\
EGNN-atom           & 20.24   & 2211.86 \\
EGNN-hierarchical           & 27.74   & 2209.14 \\
SchNet-fragment         & 14.77   & 819.17 \\
SchNet-atom         & 17.57   & 2982.90 \\
SchNet-hierarchical         & 23.33   & 2861.80 \\
TorchMD-fragment        & 20.13   & 897.87 \\
TorchMD-atom        & 24.21   & 5838.56 \\
TorchMD-hierarchical        & 33.92   & 5835.98 \\
LEFTNet-fragment        & 24.25   & 1573.83 \\
LEFTNet-atom        & 32.86   & 13592.21 \\
LEFTNet-hierarchical        & 32.97  & 13592.21 \\
GemNet-fragment         & 69.02   & 20041.35 \\
GemNet-atom         & 448.26   & 15087.50 \\
GemNet-hierarchical         & 633.70   & 13972.60 \\
DimeNet-fragment      & 53.95   & 5371.92 \\
DimeNet-atom      & 77.98   & 29456.64 \\
DimeNet-hierarchical      & 80.89   & 27150.04 \\
Equiformer-fragment     & 123.96   & 34447.65 \\
Equiformer-atom     & 873.22   & 17780.71 \\
Equiformer-hierarchical     & 881.35   & 17970.31 \\
MACE-fragment           & 59.85   & 2243.75 \\
MACE-atom           & 91.36   & 23564.17 \\
MACE-hierarchical           & 85.24   & 23709.81 \\
GET-fragment            & 59.08   & 4523.77 \\
GET-atom            & 61.68   & 5361.85 \\
GET-hierarchical            & 118.14   & 6691.07 \\
HEGNN-fragment          & 27.93   & 643.68 \\
HEGNN-atom          & 32.27   & 2298.35 \\
HEGNN-hierarchical          & 41.56   & 2300.65 \\
\midrule
ECBind-nptn    & 9.13   & 6724.58 \\
ECBind-stdt    & 9.16   & 6724.58 \\
ECBind-ptn     & 11.48   & 9302.74 \\
\bottomrule
\end{tabular}
\end{table}

\end{document}